\documentclass[11pt]{article}
\usepackage[table]{xcolor}
\usepackage[final]{acl}
\usepackage{times}
\usepackage{latexsym}
\usepackage[T1]{fontenc}
\usepackage[utf8]{inputenc}
\usepackage{microtype}
\IfFileExists{inconsolata.sty}{\usepackage{inconsolata}}{}
\usepackage{amsmath,amssymb}
\usepackage{graphicx}
\usepackage{booktabs,longtable,array,tabularx}
\usepackage{placeins}
\usepackage{listings}
\usepackage{enumitem}
\IfFileExists{xurl.sty}{\usepackage{xurl}}{}
\definecolor{engram}{HTML}{4453C4}
\definecolor{codebg}{HTML}{F5F7FA}
\hypersetup{linkcolor=engram,citecolor=engram,urlcolor=engram,
  pdftitle={When Does Selection Replace Extraction? A Pre-Registered Test of Agent Memory with a Typed Decision Model},
  pdfauthor={Rishabh Sharma}}
\usepackage[nameinlink,noabbrev]{cleveref}
\crefname{equation}{Eq.}{Eqs.}\Crefname{equation}{Eq.}{Eqs.}
\newcommand{\secfmt}[1]{%
  \crefformat{#1}{##2\S##1##3}\Crefformat{#1}{##2\S##1##3}%
  \crefrangeformat{#1}{\S\S##3##1##4--##5##2##6}\Crefrangeformat{#1}{\S\S##3##1##4--##5##2##6}%
  \crefmultiformat{#1}{\S\S##2##1##3}{ and~##2##1##3}{, ##2##1##3}{ and~##2##1##3}%
  \Crefmultiformat{#1}{\S\S##2##1##3}{ and~##2##1##3}{, ##2##1##3}{ and~##2##1##3}}
\secfmt{section}\secfmt{subsection}
\setlist{itemsep=1pt,topsep=2pt,leftmargin=*}
\newcommand{\src}[1]{}

\title{When Does Selection Replace Extraction? A Pre-Registered Test of Agent Memory with a Typed Decision Model}
\author{Rishabh Sharma\thanks{Preprint,
  \href{https://doi.org/10.5281/zenodo.22985242}{doi:10.5281/zenodo.22985242}. Pre-registered plan:
  \href{https://doi.org/10.5281/zenodo.22970745}{10.5281/zenodo.22970745}; amendment and pre-written outcome
  paragraphs: \href{https://doi.org/10.5281/zenodo.22977848}{10.5281/zenodo.22977848}.} \\
  Independent Researcher \\
  \texttt{rishabh.sharma1103@gmail.com} \And
  Rishika Lall \\
  Independent Researcher \\
  \texttt{lallrishika@gmail.com}}
\begin{document}
\maketitle
\begin{abstract}
Does conversational memory need LLM-extracted facts, or is selecting the right raw turns enough? Published results disagree. Extraction-based systems report gains from distilled facts. Recent studies find raw history with good ranking does as well, but disagree about whether ranking matters. We ran a pre-registered study on held-out LoCoMo conversations and LongMemEval. At a tight budget on LoCoMo, raw turns selected by a single call to Jev, a typed decision model, are non-inferior to an LLM-extraction memory (one-sided 95\% bound $-$3.0\src{bench/results/v3/batch\_b\_report.json\#H1/lower\_bound\_95\_one\_sided} points against a $-$5\src{docs/V3\_PLAN.md §1, §2, §4}-point margin). Blind human grading narrows the margin but does not change the result. Raw turns cost 3,061\src{bench/results/v3/batch\_b\_report.json\#write\_cost\_ratio\_engram\_over\_t0r/ratio}$\times$ less to write, and the result holds with a second answer model. Within this study, reranking's gain shrinks as the budget grows. It adds 17.4\src{bench/results/v3/batch\_a\_report.json\#fresh (T0R k3 - L0 k3)} points on LoCoMo and 9.1\src{bench/results/v3/batch\_c\_report.json\#expansion (T0R 3 - L0 3)} on LongMemEval when three of 30\src{docs/V3\_PLAN.md §1, §2, §4} candidates are kept. At generous budgets it adds 1.5\src{bench/results/v3/batch\_a\_report.json\#fresh (T0R k20 - L0 k20)} and 1.1\src{bench/results/v3/batch\_c\_report.json\#expansion (T0R 20 - L0 20)}, and extraction systems are more accurate. This suggests why published results disagree. At matched context, Jev selects as accurately as an LLM reranker (non-inferiority bound $-$2.0\src{bench/results/v3/batch\_b\_report.json\#S4/lower\_bound\_95\_one\_sided}) at a third of the latency, and more accurately than a multi-call graph traversal. Reranking lowers correct abstention. Plans, code and graded answers are released.

\end{abstract}
\section{Introduction}\label{sec:1}
Does conversational memory need LLM-extracted facts? The literature is split. Extraction-based systems report gains from distilling conversations before retrieval: mem0 \citep{chhikara2025mem0} extracts facts from each message, the LongMemEval design study \citep{wu2025longmemeval} finds that expanding index keys with extracted facts helps retrieval, and SeCom \citep{pan2025secom} segments sessions and compresses the segments before retrieval. Recent studies find that raw history, ranked well, does as well or better: SmartSearch \citep{derehag2026smartsearch} retrieves from raw history with a deterministic pipeline and a learned ranking stage, and Fidelity Before Structure \citep{an2026fidelity} finds verbatim chunks ahead of LLM-extracted artifacts in a controlled comparison. These two also disagree with each other: SmartSearch identifies ranking as the bottleneck, while Fidelity finds that reranking adds little.

We propose that the context budget, the number of retrieved items the answer model reads, accounts for part of the disagreement. When the budget keeps a few of many candidates, the choice of items decides the answer. Selection then matters, and a good selector over raw turns can stand in for extraction. When the budget is generous, similarity order already includes most of the evidence. Ranking then adds little, and extracted facts, which are more compact, are more accurate. We test the first half of this under a pre-registered plan, on conversations never used for development. Are raw turns with a single reranking call non-inferior to a strong extraction-based memory at a tight, matched budget? And how does the rerank's value change as the budget grows? Non-inferiority means we test whether raw turns are at most 5\src{docs/V3\_PLAN.md §1, §2, §4} points worse, rather than whether the two systems differ at all.

The selector is Jev, TypeSafe's typed decision model \citep{typesafe2026launch,typesafe2026jev}. It answers a fixed-option question with a probability in one short request. The extraction system is engram v2. engram is a memory system we built and described in an earlier preprint \citep{sharma2026typed}: an LLM extracts facts, and a typed decision model makes every later decision about them. engram v2 is the version used here; it extracts facts with gpt-4o-mini and types, relates and updates them with Jev. It was the most accurate system on our development conversation, and we chose it as the comparator because the test could fail against it. Choosing our own extraction system as the comparator gave us every reason to make it strong. engram v2's read path is Turns + Jev's read path over extracted facts instead of raw turns. H1 therefore holds the selector fixed and varies only what is stored: a controlled comparison of extraction and raw turns, in the spirit of Fidelity Before Structure's.

Our contributions:

\begin{enumerate}
\item \textbf{Within this study, reranking's gain over similarity search shrinks as the budget grows}: from +17.4\src{bench/results/v3/batch\_a\_report.json\#fresh (T0R k3 - L0 k3)} to +1.5\src{bench/results/v3/batch\_a\_report.json\#fresh (T0R k20 - L0 k20)} points on LoCoMo and from +9.1\src{bench/results/v3/batch\_c\_report.json\#expansion (T0R 3 - L0 3)} to +1.1\src{bench/results/v3/batch\_c\_report.json\#expansion (T0R 20 - L0 20)} on LongMemEval, as the answer model reads three and then twenty of 30\src{docs/V3\_PLAN.md §1, §2, §4} candidates. This suggests an explanation for why SmartSearch finds ranking decisive and Fidelity Before Structure finds it marginal, which we offer as an interpretation (\cref{sec:5.3}, \cref{sec:6}). \citet{kang2026retain} give complementary evidence that memory design choices depend on the budget.
\item \textbf{To our knowledge, the first pre-registered non-inferiority test in conversational-memory evaluation}, on held-out conversations, with a second answer model and blind human grading. It bounds what extraction adds at a tight budget. Under the worst grading we applied, extraction adds at most 4.7\src{bench/results/v3/human\_audit/audit\_report.json\#lenient/H1\_with\_human\_grades/lower\_bound\_95\_one\_sided (negated)} points. Human grading puts the difference at $-$1.7\src{bench/results/v3/human\_audit/audit\_report.json\#strict/H1\_with\_human\_grades/d\_bar} to $-$2.6\src{bench/results/v3/human\_audit/audit\_report.json\#lenient/H1\_with\_human\_grades/d\_bar} points (\cref{sec:5.1}, \cref{sec:5.4}). LazyMem \citep{yu2026lazymem} also prespecifies its automatic metric and uses a human audit as a sensitivity analysis; our study registers a plan, a primary test and a non-inferiority margin before any held-out run.
\item \textbf{To our knowledge, the first answer-level, matched-context evaluation of a typed decision model as the selector.} At matched context, Jev is non-inferior to a gpt-4o-mini listwise reranker (S4, lower bound $-$2.0\src{bench/results/v3/batch\_b\_report.json\#S4/lower\_bound\_95\_one\_sided}) at about a third of the latency. It is also more accurate than Jev-Mem's multi-call Jev graph traversal at matched context (S2) (\cref{sec:5.2}, \cref{sec:5.7}). This is consistent with MemReranker \citep{li2026memreranker}, where a small reranker matches gpt-4o-mini on key retrieval metrics.
\item \textbf{Diagnostics.} A per-category decomposition of where the selection ceiling comes from: shortlist misses (11.5\%\src{bench/results/v3/shortlist\_recall.json\#all\_nine/all (1 - shortlist\_recall\_any)} of questions) and rerank drops (9.2\%\src{bench/results/v3/shortlist\_recall.json\#all\_nine/all (shortlist\_recall\_any x rerank\_drops\_all\_shortlisted\_evidence)}), with temporal evidence dropped most at the rerank (\cref{sec:5.6}). And a finding about evaluation: the LLM judge's leniency interacts with answer length, so judge--human agreement differs by system (\cref{sec:5.4}, \cref{sec:6}). Judge leniency itself is documented elsewhere \citep{ren2026memlens,penfield2026locomo}; the interaction with answer length is what we add.
\end{enumerate}
What is not new: raw turns plus a reranker is a known pattern \citep{derehag2026smartsearch,nanomemory2026}, and engram v2 is a version of our earlier system \citep{sharma2026typed}. The contribution is the test, the within-study budget result and the typed selector, not a new architecture.

\section{Related Work}\label{sec:2}
\textbf{Raw history against extraction.} SmartSearch \citep{derehag2026smartsearch} argues that neither LLM structuring at ingestion nor learned retrieval policies are necessary, and ranks raw history with a CrossEncoder and ColBERT fusion stage. Fidelity Before Structure \citep{an2026fidelity} swaps only the stored representation inside one pipeline, with gpt-4o answering and a gpt-4o-mini judge giving binary grades. Verbatim chunks lead LLM-extracted artifacts by 15.9\src{cite:an2026fidelity (checked against the paper's text; paper\_v3/bib\_verification.md)} points on LoCoMo (categories 1--3, 699\src{cite:an2026fidelity (checked against the paper's text; paper\_v3/bib\_verification.md)} questions). They lead by 22.0\src{cite:an2026fidelity (checked against the paper's text; paper\_v3/bib\_verification.md)} points on LongMemEval-S (500\src{cite:an2026fidelity (checked against the paper's text; paper\_v3/bib\_verification.md)} questions). In an external-system anchor (their Appendix D), the official Mem0 package also trails verbatim chunks. With a gpt-4o-mini answerer it scores 36.6\src{cite:an2026fidelity (checked against the paper's text; paper\_v3/bib\_verification.md)}\% against 47.9\src{cite:an2026fidelity (checked against the paper's text; paper\_v3/bib\_verification.md)}\% (categories 1--3). With gpt-4o it scores 54.7\src{cite:an2026fidelity (checked against the paper's text; paper\_v3/bib\_verification.md)}\% against 69.9\src{cite:an2026fidelity (checked against the paper's text; paper\_v3/bib\_verification.md)}\% (1,540\src{cite:an2026fidelity (checked against the paper's text; paper\_v3/bib\_verification.md)} questions, categories 1--4). Nano-Memory \citep{nanomemory2026} answers from raw turns with retrieval and generation alone; EMem \citep{zhou2025emem} builds a strong baseline from near-verbatim discourse units; \citet{zeng2024structural} sweep chunks, triples, facts and summaries and find chunk-based and mixed stores strongest on LoCoMo; the LongMemEval design study \citep{wu2025longmemeval} finds round-level storage best and fact-augmented index keys helpful; and Letta reports 74.0\src{cite:letta2025filesystem (checked against the paper's text; paper\_v3/bib\_verification.md)}\% on LoCoMo for a gpt-4o-mini agent that stores conversation history in files, with no judge stated \citep{letta2025filesystem}. Zero-Mem \citep{xiao2026zeromem} removes LLM calls from every memory operation: it keeps raw traces and retrieves with BM25, dense embeddings and an entity graph built by a non-generative NER model. It is fully deterministic, whereas our selector is a typed decision model, and we test selection against extraction under pre-registration; it reports F1, so its numbers are not comparable with ours. LazyMem \citep{yu2026lazymem} defers memory construction to query time: a trained model retains and compresses only the query-relevant content of a broad retrieved pool. It rewrites text at read time, while our selector only chooses among raw turns. Our result agrees with this lineage at tight budgets and is smaller and more cautious than Fidelity's gap, as expected for an extraction system that keeps source quotes. We extend it with a registered non-inferiority margin, held-out conversations, a budget analysis and per-category recall.

\textbf{Budgets and compression.} \citet{kang2026retain} study a complementary budget-dependent decision: given identified evidence, whether to retain raw records or replace them with generated consolidations under a fixed answer-time budget. Consolidation helps when the budget is too small for the relevant raw evidence (up to 48\src{cite:kang2026retain (checked against the paper's text; paper\_v3/bib\_verification.md)} points on LongMemEval at 32\src{cite:kang2026retain (checked against the paper's text; paper\_v3/bib\_verification.md)} tokens), and retention is preferable once it fits. We study end-to-end selection instead: whether query-time ranking of raw turns can substitute for write-time extraction when only a fraction of the retrieved candidates reaches the answer model. The two sets of results are consistent. Our tight budgets (129\src{bench/results/v3/batch\_b\_report.json\#fresh/mem0 k3/tokens\_mean}--265\src{bench/results/v3/batch\_b\_report.json\#fresh/T0R k6 (matched to engram v2)/tokens\_mean} tokens) already fit several raw turns, the regime where Kang et al. also find retention competitive. Their intervention acts after evidence has been identified, while ours tests whether selection itself removes the need for extraction. The extraction advantage we observe at generous budgets is partly our selector's read-path ceiling (\cref{sec:5.3}), and does not contradict their finding. EMBER \citep{li2026ember} learns which verbatim evidence to retain under a fixed pre-query token budget, and a controlled comparison of memory substrates finds that none dominates across operating regimes \citep{huang2026harness}.

\textbf{Extraction systems.} mem0 \citep{chhikara2025mem0} extracts facts per message; we test mem0 OSS 2.1.0, and newer mem0 releases report higher, self-reported numbers \citep{mem02026state}. Graphiti/Zep \citep{rasmussen2025zep}, A-MEM \citep{xu2025amem}, MemGPT/Letta \citep{packer2023memgpt}, EverMemOS \citep{evermemos2026} and Memora \citep{memora2026} structure memory with LLM calls at write time. Several recent systems make extraction cheaper: SimpleMem \citep{liu2026simplemem} compresses interactions into compact indexed memory units, LightMem \citep{fang2025lightmem} filters and groups content in stages and consolidates offline, and LeanMem \citep{liao2026leanmem} stores each kind of content as profile, event or source-grounded record memory.

\textbf{Typed decisions in memory.} Jev-Mem \citep{jiang2026jevmem} was the first memory system built on Jev; it uses typed questions for typing, relations, routing, traversal and stopping over a multi-graph store. The AtMem--Jev article \citep{taghia2026atmem} reports that Jev reranking raises ranking metrics. We measure a Jev reranker at the answer level, against an LLM reranker and against Jev-Mem at matched context.

\textbf{Reranking in conversational memory.} SmartSearch finds ranking to be the bottleneck; Fidelity finds reranking marginal. Training-Free Lexical--Dense Fusion \citep{lexdense2026} reports an off-the-shelf cross-encoder lowering Hit@1 on conversational queries, and ConvMemory v2 \citep{convmemory2026} reports gains from a cross-encoder fine-tuned for conversation. MemReranker \citep{li2026memreranker}, a small reasoning-aware reranker for agent memory, matches gpt-4o-mini on key retrieval metrics, consistent with our finding that a typed decision model selects as accurately as an LLM reranker (S4). EARM \citep{feng2026earm} treats LLM reranking as a per-query cost and amortizes it by reusing past relevance scores; the same cost argument motivates a selector that answers in one short request. Our budget analysis offers one way these findings fit together (\cref{sec:6}).

\textbf{Evaluation validity.} Held-out conversation splits of LoCoMo already exist \citep{yan2025split,useraware2026}; our design adds pre-registration and a non-inferiority margin. Same Ranking, Different Winner \citep{samerank2026} shows that retrieval credit depends on the stored form; we score shortlist recall on raw turns only. Fidelity reports judge--human agreement of $\kappa$ = 0.897\src{cite:an2026fidelity (checked against the paper's text; paper\_v3/bib\_verification.md)} on 100\src{cite:an2026fidelity (checked against the paper's text; paper\_v3/bib\_verification.md)} questions, similar for short and long answers, with a judge instructed to be strict (their Appendix J.4); with mem0's lenient LoCoMo judge, we found that agreement depended on the system's answer style (\cref{sec:5.4}, \cref{sec:6}).

\section{Systems}\label{sec:3}
\subsection{Setup and notation}\label{sec:3.1}
A conversation is a sequence of turns $x_1, \ldots, x_T$, each with its session date. A memory system has a write function $W$ that builds a store, a read function that selects part of it for a question $q$, and an answer model $L$:

\begin{align}
& M = W(x_1, \ldots, x_T), \qquad S(q) \subseteq M, \nonumber\\
& a = L\bigl(q, \operatorname{render}(S(q))\bigr) \label{eq:memory}
\end{align}
For Turns + Jev and Turns + cosine, $M$ is the turns themselves, each stored with its date. For engram v2, $M$ is a set of facts that an LLM extracted, each with a source quote and a validity window. For Jev-Mem, $M$ is a graph whose nodes are turns and whose edges Jev types.

A typed question $Q$ has a fixed option set $O_Q$. Given a state $s$, Jev returns a probability for every option in one request, and a decision is the most probable option with that probability as its confidence:

\begin{align}
& p(o \mid s, Q), \quad o \in O_Q, \nonumber\\
& d(s, Q) = \arg\max_{o \in O_Q} p(o \mid s, Q), \nonumber\\
& \pi(s, Q) = \max_{o \in O_Q} p(o \mid s, Q) \label{eq:jev}
\end{align}
Jev does not generate text. It scores a closed set of options, so its output needs no parsing, and its confidence is a probability that can be thresholded.

Reading starts from a cosine shortlist of the $n$ stored items closest to the question, with $n$ = 30\src{docs/V3\_PLAN.md §1, §2, §4} and $e(\cdot)$ the embedding:

\begin{align}
& C(q) = \operatorname*{Top}_{n}\ \cos\bigl(e(q), e(m)\bigr), \quad m \in M \label{eq:shortlist}
\end{align}
Turns + Jev asks Jev one relevance question $Q_{\text{rel}}$ about every shortlisted turn, in one request. It keeps the turns whose relevance $\rho$ exceeds $\tau$ = 0.5\src{docs/V3\_PLAN.md §1, §2, §4}, in decreasing $\rho$. The top $f$ = 10\src{docs/V3\_PLAN.md §1, §2, §4} turns of the shortlist by cosine follow them (the cosine floor), and the answer model reads the first $k$:

\begin{align}
& \rho(m, q) = p(\text{yes} \mid m, q, Q_{\text{rel}}), \nonumber\\
& R(q) = \{m \in C(q) : \rho(m, q) > \tau\}, \nonumber\\
& S_k(q) = \operatorname{first}_k\bigl(R(q) \text{ by } \rho, \nonumber\\
& \qquad\quad \text{then } \operatorname{Top}_f C(q) \setminus R(q) \text{ by cosine}\bigr) \label{eq:select}
\end{align}
Turns + cosine reads the first $k$ turns in cosine order. Below, the subscripts $J$, $\cos$ and $E$ denote Turns + Jev, Turns + cosine and engram v2. Systems are compared at matched context. With $T_A(k)$ the mean rendered tokens per question of system $A$ at $k$, and $k_B$ the comparator's own $k$ (three), Turns + Jev runs at the $k$ whose tokens are closest, ties going to the larger $k$:

\begin{align}
& k^{*} = \arg\min_{k}\ \bigl|T_{J}(k) - T_B(k_B)\bigr| \label{eq:match}
\end{align}
The rerank's gain over similarity search at the same $k$ is

\begin{align}
& \Delta(k) = \operatorname{Acc}_{J}(k) - \operatorname{Acc}_{\cos}(k) \label{eq:delta}
\end{align}
The primary test H1 compares Turns + Jev with engram v2 question by question. Let $c_i^A$ be one if system $A$'s answer to question $i$ is judged correct and zero otherwise, $d_i$ the difference Turns + Jev minus engram v2, $\bar d$ its mean and $s$ its standard deviation over the $N$ = 778\src{bench/results/v3/batch\_b\_report.json\#H1/questions} questions. Turns + Jev is non-inferior if the one-sided 95\% lower bound clears the margin $\delta$ = 5\src{docs/V3\_PLAN.md §1, §2, §4} points, with $z_{0.95}$ = 1.645:

\begin{align}
& d_i = c_i^{J} - c_i^{E}, \nonumber\\
& \bar d - z_{0.95}\, \frac{s}{\sqrt{N}} > -\delta \label{eq:primary}
\end{align}
The share of the gap between similarity search and extraction that the rerank closes, at the H1 budget, is

\begin{align}
& G = \frac{\operatorname{Acc}_{J} - \operatorname{Acc}_{\cos}}{\operatorname{Acc}_{E} - \operatorname{Acc}_{\cos}} \label{eq:gap}
\end{align}
with each system at its matched $k$. The total cost per question adds the write cost of $r_w$ turns, the turns written per question asked (4.0\src{bench/results/v3/batch\_b\_report.json\#write/engram v2/turns / 778} on the benchmark), to the read and answer costs:

\begin{align}
& C = r_w\, c_{\text{write}} + c_{\text{read}} + c_{\text{answer}} \label{eq:cost}
\end{align}
\subsection{The systems}\label{sec:3.2}
\begin{figure*}[tbp]
\centering
\includegraphics[width=\textwidth]{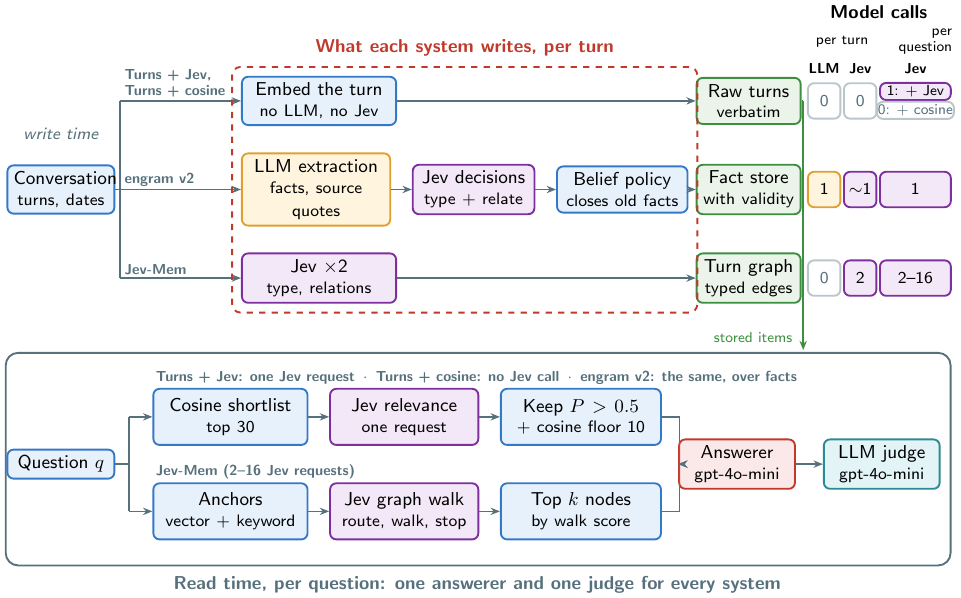}
\caption{Write path (per turn, top) and read path (per question, bottom) of Turns + Jev, Turns + cosine, engram v2 and Jev-Mem. Border colour says what does the work: code (blue), an LLM call (amber), a Jev typed decision (purple), a store (green), the answer model (red) and the judge (teal). The grid gives LLM calls and Jev requests per turn and Jev requests per question, from each system's code (Jev-Mem: its default profile); Turns + Jev and Turns + cosine share a write path and differ only per question, where Turns + Jev makes one Jev request and Turns + cosine none. A design diagram; no measured data.}\label{fig:1}
\end{figure*}

All systems use gpt-4o-mini to answer, text-embedding-3-small to embed and jev-1.13.0 for every Jev decision. Figure~\ref{fig:1} contrasts the write and read paths of Turns + Jev, engram v2 and Jev-Mem. We give the raw-turn systems descriptive names: Turns + Jev, Turns + cosine and Turns + LLM, registered as T0R, L0 and T0R-LLM in the plan. The post-hoc variant T0R-wide is Turns + Jev (wide).

\textbf{Turns + Jev.} The write path embeds each turn and stores it as ``[date] speaker: text'', with no extraction and no LLM call. The read path is \cref{eq:shortlist,eq:select}; Jev's relevance question asks whether each turn helps answer the question.

\textbf{Turns + cosine.} The same store, read in cosine order with no Jev call.

\textbf{Turns + LLM.} Turns + Jev's store and shortlist, scored by a gpt-4o-mini listwise reranker instead of Jev.

\textbf{Full context.} Every turn of the conversation, rendered as Turns + Jev renders a line, in the answer prompt.

\textbf{engram v2.} engram \citep{sharma2026typed}, our earlier system. An LLM extracts facts from each message with mem0's extraction prompt. Jev then answers typing questions and relation questions against up to ten candidate facts, and a belief policy closes superseded facts. The read path is Turns + Jev's over facts instead of turns. v2 changes two things from the preprint's version, both fixed on the development conversation before the \texttt{v2-frozen} tag. Extraction receives each message's session date, so relative dates resolve to the conversation's time. A same-attribute gate, one more Jev question per candidate, lets an update close a stored fact whose relation type differs. The v2 plan's Deviations section (\texttt{docs/\allowbreak{}V2\_\allowbreak{}PLAN.md \S{}12}) records both.

\textbf{mem0 2.1.0.} The default \texttt{add()} path: one LLM extraction call per message, with the session date as the observation date; reads are vector search.

\textbf{Jev-Mem.} Jev-Mem at commit 81574eb with its default profile and \texttt{jev\_\allowbreak{}model} pinned to jev-1.13.0, driven through its own API. Each turn is a node; each write makes two Jev requests (memory type, relations), and each read routes, traverses and stops with between two and sixteen Jev requests. Its returned turns are rendered as ``[date] speaker: text'' and answered with our prompt; its own prompts, best-of-three selection and judge are not used.

\begin{figure*}[tbp]
\centering
\includegraphics[width=\textwidth]{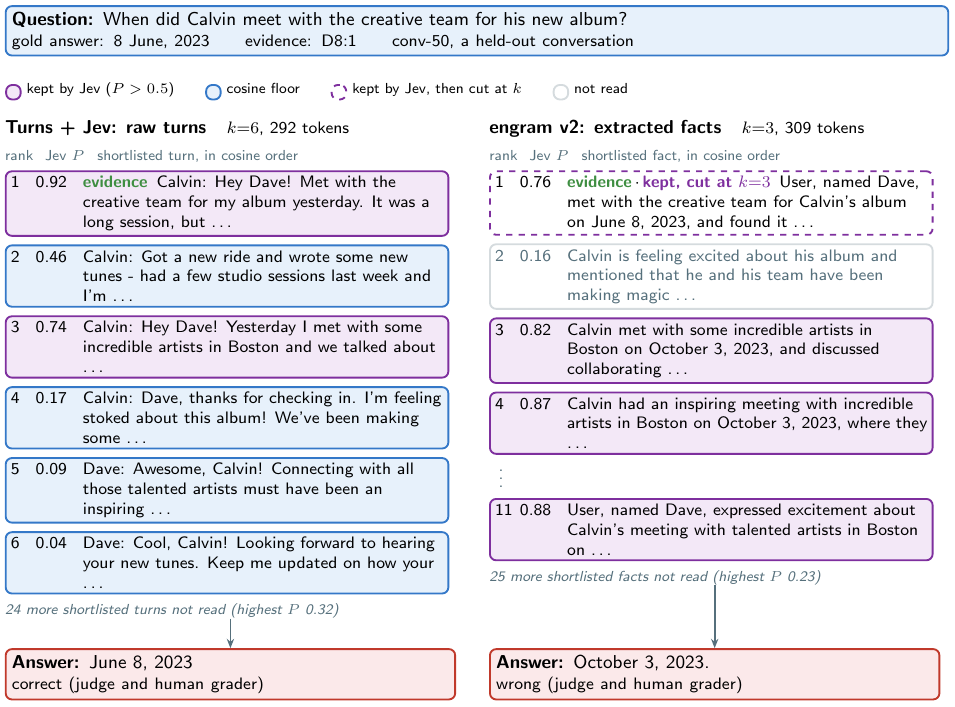}
\caption{One held-out question traced through both read paths, replayed offline from the frozen stores and the call cache (no API call; both contexts match the recorded token counts: Turns + Jev 292\src{bench/results/v3/worked\_example.json\#t0r/recorded\_tokens}, engram v2 309\src{bench/results/v3/worked\_example.json\#engram\_v2/recorded\_tokens}). Each column lists the top four of the 30\src{bench/results/v3/worked\_example.json\#t0r/shortlist (length)}-item cosine shortlist and every item the answer model read, in cosine order, with Jev's P(relevant): purple rows were kept by Jev, blue rows by the cosine floor, and the dashed row was kept by Jev but ranked below the cut at $k{=}$3. An illustration chosen by the rule in \cref{sec:3}, not evidence. \hyperref[app:J]{Appendix~J} shows a question where extraction wins, chosen by the same kind of rule.}\label{fig:2}
\end{figure*}

\textbf{A worked example.} Figure~\ref{fig:2} traces one held-out question through Turns + Jev and engram v2 at the H1 budgets. It was chosen by a fixed rule, not for effect. The question must:

\begin{enumerate}
\item be an H1 question that the judge and the human grader both scored correct for Turns + Jev and wrong for engram v2;
\item be temporal (8\src{bench/results/v3/worked\_example.json\#candidates\_meeting\_rule} questions meet the first two conditions);
\item have an evidence turn that Jev's rerank kept, not the cosine floor;
\item have replayed contexts that match the recorded ones;
\item have the shortest Turns + Jev context among those left.
\end{enumerate}
engram v2 extracted the evidence turn, but under the wrong speaker. Three other facts outranked it at $k{=}$3. Turns + Jev kept the verbatim turn with its date. \hyperref[app:J]{Appendix~J} shows the opposite case, chosen by the same kind of rule: there the evidence turn never reached Turns + Jev's shortlist, while engram v2's extracted fact did.

\section{Study Design}\label{sec:4}
\textbf{Pre-registration.} The plan was deposited before any run on the data below (10.5281/zenodo.22970745, commit b3c5dc5, tag \texttt{v3-frozen}). An amendment, with the outcome paragraphs used in \cref{sec:5.1}, followed (10.5281/zenodo.22977848, commit efae0b6, tag \texttt{v3-amended}) \citep{sharma2026v3plan,sharma2026v3amend}. It was deposited after Batch A, so the results of S1 and S2 were known when it added S7, the full LongMemEval run and the second answer model. It was deposited before any primary-test (H1) result was seen.

\textbf{Data.} LoCoMo \citep{maharana2024locomo} numbers its question categories. We name them 1 multi-hop, 2 temporal, 3 open-domain, 4 single-hop and 5 adversarial, which matches the dataset's counts over all ten conversations (282\src{bench/data/locomo10.json (count of category 1, all ten conversations)}, 321\src{bench/data/locomo10.json (count of category 2, all ten conversations)}, 96\src{bench/data/locomo10.json (count of category 3, all ten conversations)}, 841\src{bench/data/locomo10.json (count of category 4, all ten conversations)} and 446\src{bench/data/locomo10.json (count of category 5, all ten conversations)} questions). The primary data are conv-44, conv-47, conv-48, conv-49 and conv-50, never run by any system before this study. They hold 3,122\src{bench/results/v3/batch\_b\_report.json\#write/engram v2/turns} turns and 778\src{bench/results/v3/batch\_b\_report.json\#H1/questions} scored questions, plus 209\src{bench/results/v3/batch\_b\_report.json\#fresh\_adversarial/mem0 k3/questions} adversarial ones. The scored questions are 140\src{bench/results/v3/lean\_t0r\_\_heldout\_*\_\_k3.json (count of category 1)} multi-hop, 165\src{bench/results/v3/lean\_t0r\_\_heldout\_*\_\_k3.json (count of category 2)} temporal, 50\src{bench/results/v3/lean\_t0r\_\_heldout\_*\_\_k3.json (count of category 3)} open-domain and 423\src{bench/results/v3/lean\_t0r\_\_heldout\_*\_\_k3.json (count of category 4)} single-hop. conv-30, conv-41, conv-42 and conv-43 (610\src{bench/results/v3/batch\_a\_report.json\#exploratory/L0 k3/questions} scored questions), held out in an earlier study, give an exploratory replication. Development used conv-26 only. LongMemEval\_S cleaned \citep{wu2025longmemeval} provides a registered sample of 70\src{bench/results/v3/batch\_c\_report.json\#S6/questions} questions, with user turns only. The amendment added all 500\src{bench/results/v3/batch\_c\_report.json (470 scored + 30 abstention)} questions with user and assistant turns: 470\src{bench/results/v3/batch\_c\_report.json\#S7/questions} scored and 30\src{bench/results/v3/batch\_c\_report.json\#expansion/T0R 3 (abstention, correct = abstained)/questions} abstention.

\textbf{Stack.} Answers and judgments use gpt-4o-mini at temperature 0 with mem0's LoCoMo answer and judge prompts; the judge returns CORRECT or WRONG. Tokens are counted with o200k\_base over the memory block the answer model sees.

\textbf{Token matching.} Every comparison between systems holds context fixed. The comparator runs at $k{=}$3, its natural setting, and Turns + Jev runs at the matched k of \cref{eq:match}. A retrieval-only sweep over k from one to thirty gave the token counts. The sweep and the chosen k were saved before Turns + Jev answered at that k.

\textbf{Tests.} The primary test H1 is \cref{eq:primary}, with the judge's labels. The margin is half the rerank's measured effect on the development conversation. The seven secondary tests, under Holm correction at family-wise 0.05, are exact two-sided McNemar tests except S4, a non-inferiority test with the same margin: S1 Turns + Jev against Turns + cosine, S2 against Jev-Mem, S3 against mem0 and S4 against Turns + LLM on LoCoMo; S5 against mem0 and S6 against Turns + cosine on the LongMemEval sample; S7 against Turns + cosine on the full LongMemEval set. The plan's power analysis put the probability of passing H1 at 0.89\src{docs/V3\_PLAN.md §1, §2, §4} if the development difference held.

\textbf{Checks.} H1 and S1 were re-answered by Llama 3.3 70B Instruct via OpenRouter from the same contexts and judged by the same judge; a result is called model-robust only if it holds under both answer models. The first author graded every question on which the judge found exactly one of H1's two answers correct, blind to system and judge label (\cref{sec:5.4}, \hyperref[app:C]{Appendix~C}). The prespecified judge decides the test, and the human audit is a sensitivity analysis, as in LazyMem \citep{yu2026lazymem}. Shortlist recall measures where the LoCoMo evidence turns fall (\cref{sec:5.6}, \hyperref[app:D]{Appendix~D}).

\textbf{Deviations.} Every change after registration is dated in the plan and listed in \hyperref[app:B]{Appendix~B}. Apart from the amendment above, none changed a test, the margin or the planned interpretation; the mem0 serving deviation adds a caveat to S3.

\section{Results}\label{sec:5}
\subsection{Primary test (H1, registered)}\label{sec:5.1}
\begin{table*}[tbp]
\centering
\footnotesize\hyphenpenalty=10000\exhyphenpenalty=10000\setlength{\tabcolsep}{4.5pt}
\caption{H1: Turns + Jev at $k{=}$6\src{bench/results/v3/batch\_b\_report.json\#token\_match/engram v2/t0r\_k} (265\src{bench/results/v3/batch\_b\_report.json\#fresh/T0R k6 (matched to engram v2)/tokens\_mean} tokens per question) against engram v2 at $k{=}$3 (251\src{bench/results/v3/batch\_b\_report.json\#token\_match/engram v2/comparator\_k3} tokens), 778\src{bench/results/v3/batch\_b\_report.json\#H1/questions} scored questions of the five held-out conversations. The judge row is the registered test; the human rows replace the judge's labels on the 141\src{bench/results/v3/human\_audit/audit\_report.json\#strict/discordant\_questions\_graded} graded discordant questions. Differences and bounds in points.}\label{tab:1}
\begin{tabular}{@{}>{\raggedright\arraybackslash}p{95.4pt}>{\raggedleft\arraybackslash}p{32.5pt}>{\raggedleft\arraybackslash}p{37.7pt}>{\raggedleft\arraybackslash}p{53.2pt}>{\raggedleft\arraybackslash}p{47.2pt}>{\raggedleft\arraybackslash}p{54.7pt}>{\raggedright\arraybackslash}p{80.1pt}@{}}
\toprule
\textbf{Grading} & \textbf{Turns + Jev} & \textbf{engram v2} & \textbf{Difference} & \textbf{One-sided 95\% bound} & \textbf{Two-sided 95\% CI} & \textbf{Non-inferior (margin $-$5\src{docs/V3\_PLAN.md §1, §2, §4})} \\
\midrule
Judge (registered) & 77.0\%\src{bench/results/v3/batch\_b\_report.json\#H1/acc\_a} & 77.5\%\src{bench/results/v3/batch\_b\_report.json\#H1/acc\_b} & $-$0.5\src{bench/results/v3/batch\_b\_report.json\#H1/d\_bar} & $-$3.0\src{bench/results/v3/batch\_b\_report.json\#H1/lower\_bound\_95\_one\_sided} & [$-$3.5\src{bench/results/v3/batch\_b\_report.json\#H1 (d\_bar - 1.96 se)},~+2.5\src{bench/results/v3/batch\_b\_report.json\#H1 (d\_bar + 1.96 se)}] & yes \\
Human, strict & 76.3\%\src{bench/results/v3/human\_audit/audit\_report.json\#strict/H1\_with\_human\_grades/acc\_t0r} & 78.0\%\src{bench/results/v3/human\_audit/audit\_report.json\#strict/H1\_with\_human\_grades/acc\_engram} & $-$1.7\src{bench/results/v3/human\_audit/audit\_report.json\#strict/H1\_with\_human\_grades/d\_bar} & $-$3.9\src{bench/results/v3/human\_audit/audit\_report.json\#strict/H1\_with\_human\_grades/lower\_bound\_95\_one\_sided} & [$-$4.3\src{bench/results/v3/human\_audit/audit\_report.json\#strict/H1\_with\_human\_grades/ci95\_two\_sided/0},~+0.9\src{bench/results/v3/human\_audit/audit\_report.json\#strict/H1\_with\_human\_grades/ci95\_two\_sided/1}] & yes \\
Human, lenient & 77.4\%\src{bench/results/v3/human\_audit/audit\_report.json\#lenient/H1\_with\_human\_grades/acc\_t0r} & 79.9\%\src{bench/results/v3/human\_audit/audit\_report.json\#lenient/H1\_with\_human\_grades/acc\_engram} & $-$2.6\src{bench/results/v3/human\_audit/audit\_report.json\#lenient/H1\_with\_human\_grades/d\_bar} & $-$4.7\src{bench/results/v3/human\_audit/audit\_report.json\#lenient/H1\_with\_human\_grades/lower\_bound\_95\_one\_sided} & [$-$5.1\src{bench/results/v3/human\_audit/audit\_report.json\#lenient/H1\_with\_human\_grades/ci95\_two\_sided/0},~$-$0.03\src{bench/results/v3/human\_audit/audit\_report.json\#lenient/H1\_with\_human\_grades/ci95\_two\_sided/1}] & yes \\
\bottomrule
\end{tabular}
\end{table*}
The registered outcome paragraph, filled in (quoted with the plan's names; T0R is Turns + Jev):

\begin{quote}
Pass. At matched context (265\src{bench/results/v3/batch\_b\_report.json\#fresh/T0R k6 (matched to engram v2)/tokens\_mean} tokens; engram v2 251\src{bench/results/v3/batch\_b\_report.json\#token\_match/engram v2/comparator\_k3}), raw turns with a single rerank call were non-inferior to LLM-extraction memory: difference $-$0.5\src{bench/results/v3/batch\_b\_report.json\#H1/d\_bar} points, one-sided 95\% lower bound $-$3.0\src{bench/results/v3/batch\_b\_report.json\#H1/lower\_bound\_95\_one\_sided}, above the registered $-$5\src{docs/V3\_PLAN.md §1, §2, §4} margin. Whatever accuracy extraction adds at this budget is under 3.0\src{bench/results/v3/batch\_b\_report.json\#H1/lower\_bound\_95\_one\_sided (negated)} points, at 3,061\src{bench/results/v3/batch\_b\_report.json\#write\_cost\_ratio\_engram\_over\_t0r/ratio}$\times$ the write cost. The rerank closes 94\%\src{bench/results/v3/batch\_b\_report.json\#G/G} of the gap between similarity search and extraction. By category, T0R did not trail on multi-hop (74.3\src{bench/results/v3/batch\_b\_report.json\#fresh/T0R k6 (matched to engram v2)/by\_category/multi-hop} vs 72.1\src{bench/results/v3/batch\_b\_report.json\#fresh/engram v2 k3/by\_category/multi-hop}, n=140\src{bench/results/v3/lean\_t0r\_\_heldout\_*\_\_k3.json (count of category 1)}) and trailed on open-domain (56.0\src{bench/results/v3/batch\_b\_report.json\#fresh/T0R k6 (matched to engram v2)/by\_category/open-domain} vs 60.0\src{bench/results/v3/batch\_b\_report.json\#fresh/engram v2 k3/by\_category/open-domain}, n=50\src{bench/results/v3/lean\_t0r\_\_heldout\_*\_\_k3.json (count of category 3)}), contrary to what we registered for multi-hop and as we registered for open-domain.
\end{quote}

\begin{figure}[tbp]
\centering
\includegraphics[width=\columnwidth]{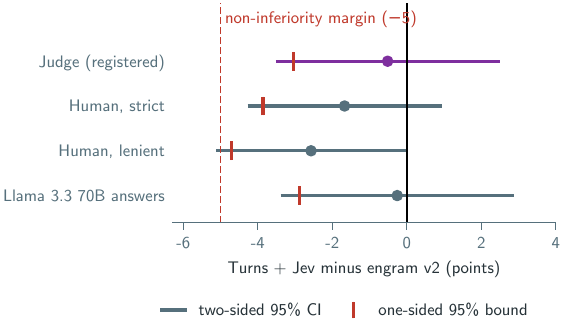}
\caption{H1 (registered) as a forest plot: Turns + Jev at $k{=}$6\src{bench/results/v3/batch\_b\_report.json\#token\_match/engram v2/t0r\_k} minus engram v2 at $k{=}$3, in points, on the 778\src{bench/results/v3/batch\_b\_report.json\#H1/questions} questions of the five held-out conversations. Bars are two-sided 95\% intervals; the red tick is the one-sided 95\% lower bound, tested against the $-$5\src{docs/V3\_PLAN.md §1, §2, §4}-point margin (dashed). The judge row is the registered test; the human rows replace the judge's labels on the 141\src{bench/results/v3/human\_audit/audit\_report.json\#strict/discordant\_questions\_graded} graded discordant questions (\cref{sec:5.4}); the Llama 3.3 70B row re-answers from the same contexts (the answer-model check of \cref{sec:4}).}\label{fig:3}
\end{figure}

The one-sided p-value is 0.0017\src{bench/results/v3/batch\_b\_report.json\#H1/p\_one\_sided}. The conversation bootstrap puts the fifth percentile of the difference at $-$2.3\src{bench/results/v3/batch\_b\_report.json\#H1/bootstrap\_conversations\_5th\_percentile} points. 69\src{bench/results/v3/batch\_b\_report.json\#H1/only\_a} questions were answered correctly only by Turns + Jev, and 73\src{bench/results/v3/batch\_b\_report.json\#H1/only\_b} only by engram v2. Human grading moves the difference to between $-$1.7\src{bench/results/v3/human\_audit/audit\_report.json\#strict/H1\_with\_human\_grades/d\_bar} and $-$2.6\src{bench/results/v3/human\_audit/audit\_report.json\#lenient/H1\_with\_human\_grades/d\_bar} points. It moves the bound to between $-$3.9\src{bench/results/v3/human\_audit/audit\_report.json\#strict/H1\_with\_human\_grades/lower\_bound\_95\_one\_sided} and $-$4.7\src{bench/results/v3/human\_audit/audit\_report.json\#lenient/H1\_with\_human\_grades/lower\_bound\_95\_one\_sided}. Under lenient grading the two-sided interval lies just below zero. By that grading engram v2 is more accurate, still inside the margin (Figure~\ref{fig:3}). This depends on keeping the judge's labels for the one ungraded question; dropping it moves the interval's upper end to +0.09\src{bench/results/v3/human\_audit/audit\_report.json\#lenient/H1\_with\_human\_grades\_ungraded\_dropped/ci95\_two\_sided/1} (\hyperref[app:C]{Appendix~C}). We therefore state the result as non-inferior within a 5\src{docs/V3\_PLAN.md §1, §2, §4}-point margin under every grading we applied. At this budget, extraction adds at most 4.7\src{bench/results/v3/human\_audit/audit\_report.json\#lenient/H1\_with\_human\_grades/lower\_bound\_95\_one\_sided (negated)} points. The category comparisons are descriptive; the categories are small and the differences are not tested.

G \cref{eq:gap} uses Turns + cosine at its own matched k (6\src{bench/results/v3/batch\_b\_report.json\#G/l0\_k}), where it scores 68.6\%\src{bench/results/v3/batch\_b\_report.json\#G/l0}. At the same token budget, one rerank call closes G = 94\%\src{bench/results/v3/batch\_b\_report.json\#G/G} of the accuracy gap between Turns + cosine and engram v2 (77.5\%\src{bench/results/v3/batch\_b\_report.json\#H1/acc\_b}). The write-cost ratio uses held-out measurements at list prices. engram v2 costs \$1.865\src{bench/results/v3/batch\_b\_report.json\#write\_cost\_ratio\_engram\_over\_t0r/engram\_v2\_per\_1k} per 1,000 turns, and Turns + Jev \$0.00061\src{bench/results/v3/batch\_b\_report.json\#write\_cost\_ratio\_engram\_over\_t0r/t0r\_per\_1k} (embeddings only).

\subsection{Secondary tests (S1--S7, registered)}\label{sec:5.2}
\begin{figure}[tbp]
\centering
\includegraphics[width=\columnwidth]{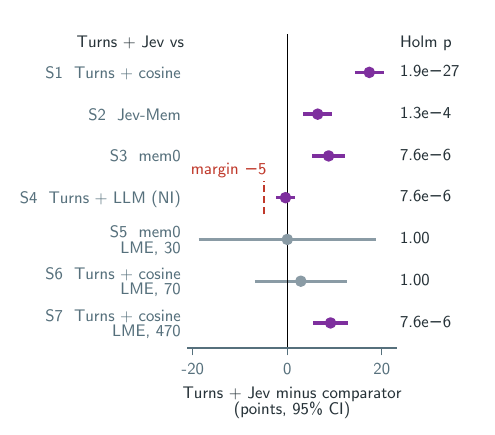}
\caption{Secondary tests S1--S7 (registered): Turns + Jev minus the comparator, in points, with paired 95\% intervals; the Holm-adjusted p is printed at the right, and purple rows are rejected after Holm correction (grey rows are not). S4 is a non-inferiority test against the $-$5\src{docs/V3\_PLAN.md §1, §2, §4}-point margin (dashed). LoCoMo tests use 778\src{bench/results/v3/batch\_b\_report.json\#H1/questions} questions; S5 and S6 use the LongMemEval sample (30\src{bench/results/v3/batch\_c\_report.json\#S5/questions} and 70\src{bench/results/v3/batch\_c\_report.json\#S6/questions} questions), S7 the full set (470\src{bench/results/v3/batch\_c\_report.json\#S7/questions}).}\label{fig:4}
\end{figure}

\begin{table*}[tbp]
\centering
\footnotesize\hyphenpenalty=10000\exhyphenpenalty=10000\setlength{\tabcolsep}{4.5pt}
\caption{Secondary tests. In each, the comparator runs at $k{=}$3 and Turns + Jev at its matched k. ``Only Turns + Jev'' and ``only other'' count questions answered correctly by one system. S4 is a non-inferiority test (one-sided p); the rest are exact two-sided McNemar tests. Holm adjustment over S1--S7. LoCoMo tests use 778\src{bench/results/v3/batch\_b\_report.json\#H1/questions} questions; LongMemEval ingestion: user turns for S5--S6, user and assistant turns for S7. Bold: rejected after Holm adjustment (family-wise 0.05).}\label{tab:2}
\begin{tabular}{@{}>{\raggedright\arraybackslash}p{33.6pt}>{\raggedright\arraybackslash}p{173.4pt}>{\raggedleft\arraybackslash}p{29.2pt}>{\raggedleft\arraybackslash}p{29.2pt}>{\raggedleft\arraybackslash}p{27.4pt}>{\raggedleft\arraybackslash}p{33.2pt}>{\raggedleft\arraybackslash}p{33.0pt}>{\raggedleft\arraybackslash}p{33.0pt}@{}}
\toprule
\textbf{Test} & \textbf{Turns + Jev vs} & \textbf{Turns + Jev k} & \textbf{Turns + Jev} & \textbf{Other} & \textbf{Only Turns + Jev / only other} & \textbf{p} & \textbf{Holm p} \\
\midrule
S1 & Turns + cosine (LoCoMo) & 3\src{bench/results/v3/batch\_a\_report.json\#token\_match/L0/t0r\_k} & 77.2\%\src{bench/results/v3/batch\_a\_report.json\#S1/acc\_a} & 59.9\%\src{bench/results/v3/batch\_a\_report.json\#S1/acc\_b} & 152\src{bench/results/v3/batch\_a\_report.json\#S1/only\_a} / 17\src{bench/results/v3/batch\_a\_report.json\#S1/only\_b} & 2.7e$-$28\src{bench/results/v3/batch\_a\_report.json\#S1/p\_two\_sided} & \textbf{1.9e$-$27\src{bench/results/v3/batch\_c\_report.json\#holm\_S1\_S7/S1/holm\_adjusted\_p}} \\
S2 & Jev-Mem (LoCoMo) & 4\src{bench/results/v3/batch\_a\_report.json\#token\_match/Jev-Mem/t0r\_k} & 77.0\%\src{bench/results/v3/batch\_a\_report.json\#S2/acc\_a} & 70.6\%\src{bench/results/v3/batch\_a\_report.json\#S2/acc\_b} & 98\src{bench/results/v3/batch\_a\_report.json\#S2/only\_a} / 48\src{bench/results/v3/batch\_a\_report.json\#S2/only\_b} & 4.3e$-$5\src{bench/results/v3/batch\_a\_report.json\#S2/p\_two\_sided} & \textbf{1.3e$-$4\src{bench/results/v3/batch\_c\_report.json\#holm\_S1\_S7/S2/holm\_adjusted\_p}} \\
S3 & mem0 (LoCoMo) & 3\src{bench/results/v3/batch\_b\_report.json\#token\_match/mem0/t0r\_k} & 77.2\%\src{bench/results/v3/batch\_b\_report.json\#S3/acc\_a} & 68.5\%\src{bench/results/v3/batch\_b\_report.json\#S3/acc\_b} & 134\src{bench/results/v3/batch\_b\_report.json\#S3/only\_a} / 66\src{bench/results/v3/batch\_b\_report.json\#S3/only\_b} & 1.7e$-$6\src{bench/results/v3/batch\_b\_report.json\#S3/p\_two\_sided} & \textbf{7.6e$-$6\src{bench/results/v3/batch\_c\_report.json\#holm\_S1\_S7/S3/holm\_adjusted\_p}} \\
S4 & Turns + LLM (LoCoMo, non-inferiority) & 3\src{bench/results/v3/batch\_b\_report.json\#token\_match/T0R-LLM/t0r\_k} & 77.2\%\src{bench/results/v3/batch\_b\_report.json\#S4/acc\_a} & 77.6\%\src{bench/results/v3/batch\_b\_report.json\#S4/acc\_b} & 28\src{bench/results/v3/batch\_b\_report.json\#S4/only\_a} / 31\src{bench/results/v3/batch\_b\_report.json\#S4/only\_b} & 1.5e$-$6\src{bench/results/v3/batch\_b\_report.json\#S4/p\_one\_sided} & \textbf{7.6e$-$6\src{bench/results/v3/batch\_c\_report.json\#holm\_S1\_S7/S4/holm\_adjusted\_p}} \\
S5 & mem0 (LongMemEval, 30\src{bench/results/v3/batch\_c\_report.json\#S5/questions} knowledge-update) & 2\src{bench/results/v3/batch\_c\_report.json\#token\_match/sample: T0R vs mem0 3/t0r\_k} & 70.0\%\src{bench/results/v3/batch\_c\_report.json\#S5/acc\_a} & 70.0\%\src{bench/results/v3/batch\_c\_report.json\#S5/acc\_b} & 4\src{bench/results/v3/batch\_c\_report.json\#S5/only\_a} / 4\src{bench/results/v3/batch\_c\_report.json\#S5/only\_b} & 1.00\src{bench/results/v3/batch\_c\_report.json\#S5/p\_two\_sided} & 1.00\src{bench/results/v3/batch\_c\_report.json\#holm\_S1\_S7/S5/holm\_adjusted\_p} \\
S6 & Turns + cosine (LongMemEval sample, 70\src{bench/results/v3/batch\_c\_report.json\#S6/questions}) & 3\src{bench/results/v3/batch\_c\_report.json\#token\_match/sample: T0R vs L0 3/t0r\_k} & 68.6\%\src{bench/results/v3/batch\_c\_report.json\#S6/acc\_a} & 65.7\%\src{bench/results/v3/batch\_c\_report.json\#S6/acc\_b} & 7\src{bench/results/v3/batch\_c\_report.json\#S6/only\_a} / 5\src{bench/results/v3/batch\_c\_report.json\#S6/only\_b} & 0.77\src{bench/results/v3/batch\_c\_report.json\#S6/p\_two\_sided} & 1.00\src{bench/results/v3/batch\_c\_report.json\#holm\_S1\_S7/S6/holm\_adjusted\_p} \\
S7 & Turns + cosine (LongMemEval, 470\src{bench/results/v3/batch\_c\_report.json\#S7/questions}) & 3\src{bench/results/v3/batch\_c\_report.json\#token\_match/expansion: T0R vs L0 3/t0r\_k} & 66.8\%\src{bench/results/v3/batch\_c\_report.json\#S7/acc\_a} & 57.7\%\src{bench/results/v3/batch\_c\_report.json\#S7/acc\_b} & 61\src{bench/results/v3/batch\_c\_report.json\#S7/only\_a} / 18\src{bench/results/v3/batch\_c\_report.json\#S7/only\_b} & 1.3e$-$6\src{bench/results/v3/batch\_c\_report.json\#S7/p\_two\_sided} & \textbf{7.6e$-$6\src{bench/results/v3/batch\_c\_report.json\#holm\_S1\_S7/S7/holm\_adjusted\_p}} \\
\bottomrule
\end{tabular}
\end{table*}
S1, S2, S3, S4 and S7 are rejected after Holm correction; S5 and S6 are not. On LoCoMo, at matched context, Turns + Jev was more accurate than similarity search (S1), Jev-Mem (S2) and mem0 (S3). It was non-inferior to the LLM reranker (S4: difference $-$0.4\src{bench/results/v3/batch\_b\_report.json\#S4/d\_bar} points, lower bound $-$2.0\src{bench/results/v3/batch\_b\_report.json\#S4/lower\_bound\_95\_one\_sided}). S3 carries a caveat: mem0's extraction was served through OpenRouter, about half of it by Azure (\hyperref[app:G]{Appendix~G}). On the LongMemEval sample, S5 detected no difference between Turns + Jev and mem0 on 30\src{bench/results/v3/batch\_c\_report.json\#S5/questions} knowledge-update questions, which is too few to establish equivalence, and S6 detected none between Turns + Jev and Turns + cosine on 70\src{bench/results/v3/batch\_c\_report.json\#S6/questions} questions. On the full set, S7 found Turns + Jev more accurate than Turns + cosine by +9.1\src{bench/results/v3/batch\_c\_report.json\#S7 (acc\_a - acc\_b)} points. By the registered rule, LongMemEval holds: S7 favours Turns + Jev after Holm correction and S5 does not favour mem0. Figure~\ref{fig:4} shows the paired differences with their intervals.

\subsection{The budget dependence of reranking}\label{sec:5.3}
\begin{figure}[tbp]
\centering
\includegraphics[width=\columnwidth]{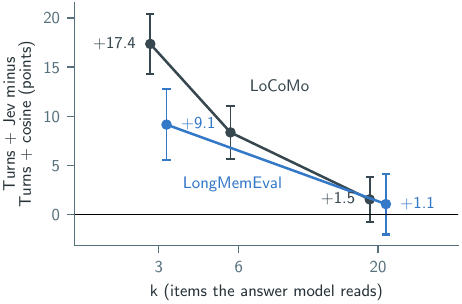}
\caption{The rerank's gain over similarity search (Turns + Jev minus Turns + cosine, paired, in points, with 95\% intervals) against k. LoCoMo: 778\src{bench/results/v3/batch\_b\_report.json\#H1/questions} questions of the five held-out conversations at $k{=}$3, $k{=}$6 and $k{=}$20; LongMemEval: 470\src{bench/results/v3/batch\_c\_report.json\#S7/questions} non-abstention questions, user and assistant turns, at $k{=}$3 and $k{=}$20. The $k{=}$3 points are registered tests (S1, S7); the others are descriptive.}\label{fig:5}
\end{figure}

The rerank's gain over similarity search, $\Delta(k)$ of \cref{eq:delta}, depends on how many candidates the budget keeps (Figure~\ref{fig:5}). On LoCoMo it is +17.4\src{bench/results/v3/batch\_a\_report.json\#fresh (T0R k3 - L0 k3)} points at $k{=}$3 and +1.5\src{bench/results/v3/batch\_a\_report.json\#fresh (T0R k20 - L0 k20)} at $k{=}$20. On the full LongMemEval set it is +9.1\src{bench/results/v3/batch\_c\_report.json\#expansion (T0R 3 - L0 3)} points at $k{=}$3 (S7) and +1.1\src{bench/results/v3/batch\_c\_report.json\#expansion (T0R 20 - L0 20)} at $k{=}$20. With three of 30\src{docs/V3\_PLAN.md §1, §2, §4} candidates kept, ordering decides which evidence reaches the answer model; with twenty kept, cosine order already includes most of it. The $k{=}$3 gains are registered tests (S1, S7); the $k{=}$20 differences are descriptive. The $k{=}$20 differences also mix the budget with Turns + Jev's own read-path ceiling. At $k{=}$20, Turns + Jev reads 496\src{bench/results/v3/batch\_a\_report.json\#fresh/T0R k20/tokens\_mean} tokens against Turns + cosine's 826\src{bench/results/v3/batch\_a\_report.json\#fresh/L0 k20/tokens\_mean}. It keeps only turns scored above 0.5\src{docs/V3\_PLAN.md §1, §2, §4}, plus the 10\src{docs/V3\_PLAN.md §1, §2, §4}-turn cosine floor, so it often cannot fill twenty slots. The post-hoc Turns + Jev (wide), which keeps the top k with no cut-off, scored 81.5\%\src{bench/results/v3\_posthoc/report.json\#comparison/T0R-wide 47 (POST-HOC EXPLORATORY (T0R-wide; docs/V3\_PLAN.md §12))/accuracy} at 2,000\src{bench/results/v3\_posthoc/report.json\#comparison/T0R-wide 47 (POST-HOC EXPLORATORY (T0R-wide; docs/V3\_PLAN.md §12))/tokens\_mean} tokens (\cref{sec:5.6}). So the decline may be less steep for a wider read path. This is a post-hoc hypothesis, not a result.

\begin{figure*}[tbp]
\centering
\includegraphics[width=\textwidth]{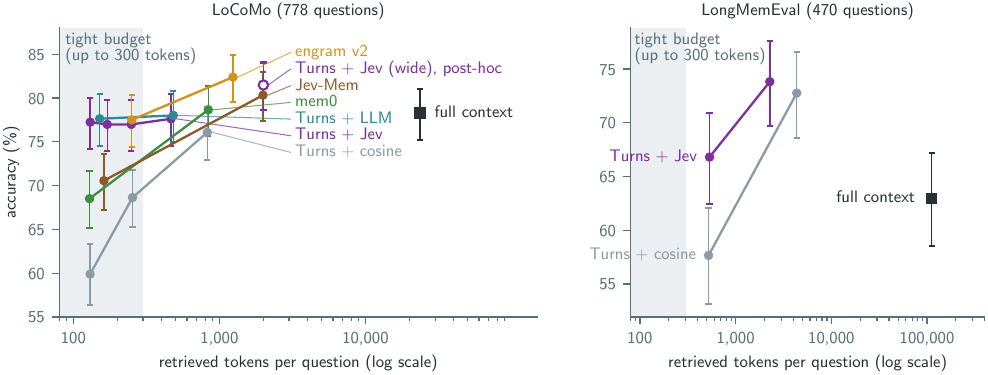}
\caption{Accuracy against retrieved tokens per question (log scale), with Wilson 95\% intervals. Left: LoCoMo, 778\src{bench/results/v3/batch\_b\_report.json\#H1/questions} questions of the five held-out conversations, each system at each k it was run. Right: LongMemEval, 470\src{bench/results/v3/batch\_c\_report.json\#S7/questions} non-abstention questions, user and assistant turns; only Turns + Jev, Turns + cosine and full context ran on all 500\src{bench/results/v3/batch\_c\_report.json (470 scored + 30 abstention)} LongMemEval questions (mem0 ran only on the 30\src{bench/results/v3/batch\_c\_report.json\#S5/questions} knowledge-update questions of the registered sample, and engram v2 and Jev-Mem not at all), which is why the right panel has three systems. Shaded: the tight budget (at most 300\src{figure setting: the tight-budget region shaded in the accuracy-context figure (paper\_v3/figures.py)} tokens). the hollow point, Turns + Jev (wide), is post-hoc; the other points are registered runs, compared descriptively except in the tests of \cref{sec:5.1}--\cref{sec:5.3}.}\label{fig:6}
\end{figure*}

Turns + Jev at $k{=}$3 is within 1.0\src{bench/results/v3/batch\_b\_report.json\#fresh (full context - T0R k3)} points of full context on LoCoMo while reading 139\src{bench/results/v3/batch\_a\_report.json\#fresh/T0R k3/tokens\_mean} tokens per question instead of 23,631\src{bench/results/v3/batch\_b\_report.json\#fresh/full context all turns/tokens\_mean}. At generous budgets the ordering reverses (Table~\ref{tab:3}, Figure~\ref{fig:6}). engram v2 at $k{=}$20 was the most accurate system we measured: 82.4\%\src{bench/results/v3/batch\_b\_report.json\#fresh/engram v2 k20/accuracy} with 1,238\src{bench/results/v3/batch\_b\_report.json\#fresh/engram v2 k20/tokens\_mean} tokens. Jev-Mem at $k{=}$40 followed, with 80.3\%\src{bench/results/v3/batch\_a\_report.json\#fresh/Jev-Mem k40/accuracy} at 1,987\src{bench/results/v3/batch\_a\_report.json\#fresh/Jev-Mem k40/tokens\_mean} tokens. mem0 at $k{=}$20 scored 78.7\%\src{bench/results/v3/batch\_b\_report.json\#fresh/mem0 k20/accuracy} with 839\src{bench/results/v3/batch\_b\_report.json\#fresh/mem0 k20/tokens\_mean} tokens. Full context scored 78.3\%\src{bench/results/v3/batch\_b\_report.json\#fresh/full context all turns/accuracy}. Turns + Jev stays near 77.6\%\src{bench/results/v3/batch\_a\_report.json\#fresh/T0R k20/accuracy} at any k (\cref{sec:5.6}). This comparison is descriptive, not a registered test, and the systems are not token-matched. In particular, Jev-Mem at its default $k{=}$40 is more accurate than Turns + Jev's ceiling (80.3\%\src{bench/results/v3/batch\_a\_report.json\#fresh/Jev-Mem k40/accuracy} against 77.6\%\src{bench/results/v3/batch\_a\_report.json\#fresh/T0R k20/accuracy}); Turns + Jev beats it only at matched context (S2).

\begin{table*}[tbp]
\centering
\footnotesize\hyphenpenalty=10000\exhyphenpenalty=10000\setlength{\tabcolsep}{3.5pt}
\caption{LoCoMo, five held-out conversations, 778\src{bench/results/v3/batch\_b\_report.json\#H1/questions} scored questions: accuracy (\%), tokens per question, write cost per 1,000 turns and read cost per query at list prices (as in Table~\ref{tab:5}: the read cost is the Jev or LLM rerank call, excluding the answer call; $\approx$0$^\dagger$: embedding only, the read path makes no model call, only a query embedding, which is not priced; ``--'': none), and accuracy by category. Rows are grouped by budget. Bold: best in column within the budget group (highest accuracy, lowest write cost); read costs are not bolded, because the lowest are the unpriced embedding-only reads.}\label{tab:3}
\begin{tabular}{@{}>{\raggedright\arraybackslash}p{83.2pt}>{\raggedleft\arraybackslash}p{40.6pt}>{\raggedleft\arraybackslash}p{31.6pt}>{\raggedleft\arraybackslash}p{30.8pt}>{\raggedleft\arraybackslash}p{35.3pt}>{\raggedleft\arraybackslash}p{38.2pt}>{\raggedleft\arraybackslash}p{41.5pt}>{\raggedleft\arraybackslash}p{53.5pt}>{\raggedleft\arraybackslash}p{44.5pt}@{}}
\toprule
\textbf{System, setting} & \textbf{Accuracy} & \textbf{Tokens} & \textbf{Write \$/1k} & \textbf{Read \$/query} & \textbf{Multi-hop} & \textbf{Temporal} & \textbf{Open-domain} & \textbf{Single-hop} \\
\midrule
\multicolumn{9}{@{}l}{\emph{Tight budget ($k{=}$3 or $k{=}$6)}} \\
Turns + cosine, $k{=}$3 & 59.9\%\src{bench/results/v3/batch\_a\_report.json\#fresh/L0 k3/accuracy} & 130\src{bench/results/v3/batch\_a\_report.json\#fresh/L0 k3/tokens\_mean} & \textbf{\$0.0006\src{bench/results/v3/batch\_a\_report.json\#write/L0 / T0R (one store)/write\_cost\_per\_1k/embeddings}} & $\approx$0$^\dagger$ & 54.3\src{bench/results/v3/batch\_a\_report.json\#fresh/L0 k3/by\_category/multi-hop} & 55.2\src{bench/results/v3/batch\_a\_report.json\#fresh/L0 k3/by\_category/temporal} & 42.0\src{bench/results/v3/batch\_a\_report.json\#fresh/L0 k3/by\_category/open-domain} & 65.7\src{bench/results/v3/batch\_a\_report.json\#fresh/L0 k3/by\_category/single-hop} \\
Turns + cosine, $k{=}$6 & 68.6\%\src{bench/results/v3/batch\_b\_report.json\#fresh/L0 k6 (matched to engram v2, for G)/accuracy} & 254\src{bench/results/v3/batch\_b\_report.json\#fresh/L0 k6 (matched to engram v2, for G)/tokens\_mean} & \textbf{\$0.0006\src{bench/results/v3/batch\_a\_report.json\#write/L0 / T0R (one store)/write\_cost\_per\_1k/embeddings}} & $\approx$0$^\dagger$ & 61.4\src{bench/results/v3/batch\_b\_report.json\#fresh/L0 k6 (matched to engram v2, for G)/by\_category/multi-hop} & 60.6\src{bench/results/v3/batch\_b\_report.json\#fresh/L0 k6 (matched to engram v2, for G)/by\_category/temporal} & 54.0\src{bench/results/v3/batch\_b\_report.json\#fresh/L0 k6 (matched to engram v2, for G)/by\_category/open-domain} & 75.9\src{bench/results/v3/batch\_b\_report.json\#fresh/L0 k6 (matched to engram v2, for G)/by\_category/single-hop} \\
Turns + Jev, $k{=}$3 & 77.2\%\src{bench/results/v3/batch\_a\_report.json\#fresh/T0R k3/accuracy} & 139\src{bench/results/v3/batch\_a\_report.json\#fresh/T0R k3/tokens\_mean} & \textbf{\$0.0006\src{bench/results/v3/batch\_a\_report.json\#write/L0 / T0R (one store)/write\_cost\_per\_1k/embeddings}} & \$0.00020\src{bench/results/v3/batch\_a\_report.json\#fresh/T0R k3/read\_cost\_per\_query} & \textbf{75.7\src{bench/results/v3/batch\_a\_report.json\#fresh/T0R k3/by\_category/multi-hop}} & 69.1\src{bench/results/v3/batch\_a\_report.json\#fresh/T0R k3/by\_category/temporal} & 58.0\src{bench/results/v3/batch\_a\_report.json\#fresh/T0R k3/by\_category/open-domain} & \textbf{83.2\src{bench/results/v3/batch\_a\_report.json\#fresh/T0R k3/by\_category/single-hop}} \\
Turns + Jev, $k{=}$6 & 77.0\%\src{bench/results/v3/batch\_b\_report.json\#fresh/T0R k6 (matched to engram v2)/accuracy} & 265\src{bench/results/v3/batch\_b\_report.json\#fresh/T0R k6 (matched to engram v2)/tokens\_mean} & \textbf{\$0.0006\src{bench/results/v3/batch\_a\_report.json\#write/L0 / T0R (one store)/write\_cost\_per\_1k/embeddings}} & \$0.00020\src{bench/results/v3/batch\_b\_report.json\#fresh/T0R k6 (matched to engram v2)/read\_cost\_per\_query} & 74.3\src{bench/results/v3/batch\_b\_report.json\#fresh/T0R k6 (matched to engram v2)/by\_category/multi-hop} & 72.1\src{bench/results/v3/batch\_b\_report.json\#fresh/T0R k6 (matched to engram v2)/by\_category/temporal} & 56.0\src{bench/results/v3/batch\_b\_report.json\#fresh/T0R k6 (matched to engram v2)/by\_category/open-domain} & 82.3\src{bench/results/v3/batch\_b\_report.json\#fresh/T0R k6 (matched to engram v2)/by\_category/single-hop} \\
Turns + LLM, $k{=}$3 & \textbf{77.6\%\src{bench/results/v3/batch\_b\_report.json\#fresh/T0R-LLM k3/accuracy}} & 143\src{bench/results/v3/batch\_b\_report.json\#fresh/T0R-LLM k3/tokens\_mean} & \textbf{\$0.0006\src{bench/results/v3/batch\_a\_report.json\#write/L0 / T0R (one store)/write\_cost\_per\_1k/embeddings}} & \$0.00023\src{bench/results/v3/batch\_b\_report.json\#fresh/T0R-LLM k3/read\_cost\_per\_query} & 73.6\src{bench/results/v3/batch\_b\_report.json\#fresh/T0R-LLM k3/by\_category/multi-hop} & 72.7\src{bench/results/v3/batch\_b\_report.json\#fresh/T0R-LLM k3/by\_category/temporal} & 58.0\src{bench/results/v3/batch\_b\_report.json\#fresh/T0R-LLM k3/by\_category/open-domain} & \textbf{83.2\src{bench/results/v3/batch\_b\_report.json\#fresh/T0R-LLM k3/by\_category/single-hop}} \\
engram v2, $k{=}$3 & 77.5\%\src{bench/results/v3/batch\_b\_report.json\#fresh/engram v2 k3/accuracy} & 251\src{bench/results/v3/batch\_b\_report.json\#fresh/engram v2 k3/tokens\_mean} & \$1.865\src{bench/results/v3/batch\_b\_report.json\#write/engram v2/write\_cost\_per\_1k (sum)} & \$0.00018\src{bench/results/v3/batch\_b\_report.json\#fresh/engram v2 k3/read\_cost\_per\_query} & 72.1\src{bench/results/v3/batch\_b\_report.json\#fresh/engram v2 k3/by\_category/multi-hop} & \textbf{77.6\src{bench/results/v3/batch\_b\_report.json\#fresh/engram v2 k3/by\_category/temporal}} & \textbf{60.0\src{bench/results/v3/batch\_b\_report.json\#fresh/engram v2 k3/by\_category/open-domain}} & 81.3\src{bench/results/v3/batch\_b\_report.json\#fresh/engram v2 k3/by\_category/single-hop} \\
mem0, $k{=}$3 & 68.5\%\src{bench/results/v3/batch\_b\_report.json\#fresh/mem0 k3/accuracy} & 129\src{bench/results/v3/batch\_b\_report.json\#fresh/mem0 k3/tokens\_mean} & \$1.321\src{bench/results/v3/batch\_b\_report.json\#write/mem0/write\_cost\_per\_1k (sum)} & $\approx$0$^\dagger$ & 56.4\src{bench/results/v3/batch\_b\_report.json\#fresh/mem0 k3/by\_category/multi-hop} & 67.3\src{bench/results/v3/batch\_b\_report.json\#fresh/mem0 k3/by\_category/temporal} & 56.0\src{bench/results/v3/batch\_b\_report.json\#fresh/mem0 k3/by\_category/open-domain} & 74.5\src{bench/results/v3/batch\_b\_report.json\#fresh/mem0 k3/by\_category/single-hop} \\
Jev-Mem, $k{=}$3 & 70.6\%\src{bench/results/v3/batch\_a\_report.json\#fresh/Jev-Mem k3/accuracy} & 162\src{bench/results/v3/batch\_a\_report.json\#fresh/Jev-Mem k3/tokens\_mean} & \$0.212\src{bench/results/v3/batch\_a\_report.json\#write/Jev-Mem/write\_cost\_per\_1k (jev + embeddings)} & \$0.00120\src{bench/results/v3/batch\_a\_report.json\#fresh/Jev-Mem k3/read\_cost\_per\_query} & 57.9\src{bench/results/v3/batch\_a\_report.json\#fresh/Jev-Mem k3/by\_category/multi-hop} & 64.2\src{bench/results/v3/batch\_a\_report.json\#fresh/Jev-Mem k3/by\_category/temporal} & 50.0\src{bench/results/v3/batch\_a\_report.json\#fresh/Jev-Mem k3/by\_category/open-domain} & 79.7\src{bench/results/v3/batch\_a\_report.json\#fresh/Jev-Mem k3/by\_category/single-hop} \\
\midrule
\multicolumn{9}{@{}l}{\emph{Generous budget ($k{=}$20 or $k{=}$40, and full context)}} \\
Turns + cosine, $k{=}$20 & 76.1\%\src{bench/results/v3/batch\_a\_report.json\#fresh/L0 k20/accuracy} & 826\src{bench/results/v3/batch\_a\_report.json\#fresh/L0 k20/tokens\_mean} & \textbf{\$0.0006\src{bench/results/v3/batch\_a\_report.json\#write/L0 / T0R (one store)/write\_cost\_per\_1k/embeddings}} & $\approx$0$^\dagger$ & 68.6\src{bench/results/v3/batch\_a\_report.json\#fresh/L0 k20/by\_category/multi-hop} & 68.5\src{bench/results/v3/batch\_a\_report.json\#fresh/L0 k20/by\_category/temporal} & 58.0\src{bench/results/v3/batch\_a\_report.json\#fresh/L0 k20/by\_category/open-domain} & 83.7\src{bench/results/v3/batch\_a\_report.json\#fresh/L0 k20/by\_category/single-hop} \\
Turns + Jev, $k{=}$20 & 77.6\%\src{bench/results/v3/batch\_a\_report.json\#fresh/T0R k20/accuracy} & 496\src{bench/results/v3/batch\_a\_report.json\#fresh/T0R k20/tokens\_mean} & \textbf{\$0.0006\src{bench/results/v3/batch\_a\_report.json\#write/L0 / T0R (one store)/write\_cost\_per\_1k/embeddings}} & \$0.00020\src{bench/results/v3/batch\_a\_report.json\#fresh/T0R k20/read\_cost\_per\_query} & 75.0\src{bench/results/v3/batch\_a\_report.json\#fresh/T0R k20/by\_category/multi-hop} & 72.7\src{bench/results/v3/batch\_a\_report.json\#fresh/T0R k20/by\_category/temporal} & 58.0\src{bench/results/v3/batch\_a\_report.json\#fresh/T0R k20/by\_category/open-domain} & 82.7\src{bench/results/v3/batch\_a\_report.json\#fresh/T0R k20/by\_category/single-hop} \\
Turns + LLM, $k{=}$20 & 78.0\%\src{bench/results/v3/batch\_b\_report.json\#fresh/T0R-LLM k20/accuracy} & 456\src{bench/results/v3/batch\_b\_report.json\#fresh/T0R-LLM k20/tokens\_mean} & \textbf{\$0.0006\src{bench/results/v3/batch\_a\_report.json\#write/L0 / T0R (one store)/write\_cost\_per\_1k/embeddings}} & \$0.00023\src{bench/results/v3/batch\_b\_report.json\#fresh/T0R-LLM k20/read\_cost\_per\_query} & 72.9\src{bench/results/v3/batch\_b\_report.json\#fresh/T0R-LLM k20/by\_category/multi-hop} & 71.5\src{bench/results/v3/batch\_b\_report.json\#fresh/T0R-LLM k20/by\_category/temporal} & 60.0\src{bench/results/v3/batch\_b\_report.json\#fresh/T0R-LLM k20/by\_category/open-domain} & 84.4\src{bench/results/v3/batch\_b\_report.json\#fresh/T0R-LLM k20/by\_category/single-hop} \\
engram v2, $k{=}$20 & \textbf{82.4\%\src{bench/results/v3/batch\_b\_report.json\#fresh/engram v2 k20/accuracy}} & 1,238\src{bench/results/v3/batch\_b\_report.json\#fresh/engram v2 k20/tokens\_mean} & \$1.865\src{bench/results/v3/batch\_b\_report.json\#write/engram v2/write\_cost\_per\_1k (sum)} & \$0.00018\src{bench/results/v3/batch\_b\_report.json\#fresh/engram v2 k20/read\_cost\_per\_query} & 77.9\src{bench/results/v3/batch\_b\_report.json\#fresh/engram v2 k20/by\_category/multi-hop} & \textbf{80.0\src{bench/results/v3/batch\_b\_report.json\#fresh/engram v2 k20/by\_category/temporal}} & \textbf{64.0\src{bench/results/v3/batch\_b\_report.json\#fresh/engram v2 k20/by\_category/open-domain}} & 87.0\src{bench/results/v3/batch\_b\_report.json\#fresh/engram v2 k20/by\_category/single-hop} \\
mem0, $k{=}$20 & 78.7\%\src{bench/results/v3/batch\_b\_report.json\#fresh/mem0 k20/accuracy} & 839\src{bench/results/v3/batch\_b\_report.json\#fresh/mem0 k20/tokens\_mean} & \$1.321\src{bench/results/v3/batch\_b\_report.json\#write/mem0/write\_cost\_per\_1k (sum)} & $\approx$0$^\dagger$ & 74.3\src{bench/results/v3/batch\_b\_report.json\#fresh/mem0 k20/by\_category/multi-hop} & 76.4\src{bench/results/v3/batch\_b\_report.json\#fresh/mem0 k20/by\_category/temporal} & 54.0\src{bench/results/v3/batch\_b\_report.json\#fresh/mem0 k20/by\_category/open-domain} & 83.9\src{bench/results/v3/batch\_b\_report.json\#fresh/mem0 k20/by\_category/single-hop} \\
Jev-Mem, $k{=}$40 & 80.3\%\src{bench/results/v3/batch\_a\_report.json\#fresh/Jev-Mem k40/accuracy} & 1,987\src{bench/results/v3/batch\_a\_report.json\#fresh/Jev-Mem k40/tokens\_mean} & \$0.212\src{bench/results/v3/batch\_a\_report.json\#write/Jev-Mem/write\_cost\_per\_1k (jev + embeddings)} & \$0.00174\src{bench/results/v3/batch\_a\_report.json\#fresh/Jev-Mem k40/read\_cost\_per\_query} & 76.4\src{bench/results/v3/batch\_a\_report.json\#fresh/Jev-Mem k40/by\_category/multi-hop} & 73.9\src{bench/results/v3/batch\_a\_report.json\#fresh/Jev-Mem k40/by\_category/temporal} & 54.0\src{bench/results/v3/batch\_a\_report.json\#fresh/Jev-Mem k40/by\_category/open-domain} & 87.2\src{bench/results/v3/batch\_a\_report.json\#fresh/Jev-Mem k40/by\_category/single-hop} \\
Full context & 78.3\%\src{bench/results/v3/batch\_b\_report.json\#fresh/full context all turns/accuracy} & 23,631\src{bench/results/v3/batch\_b\_report.json\#fresh/full context all turns/tokens\_mean} & -- & -- & \textbf{78.6\src{bench/results/v3/batch\_b\_report.json\#fresh/full context all turns/by\_category/multi-hop}} & 57.0\src{bench/results/v3/batch\_b\_report.json\#fresh/full context all turns/by\_category/temporal} & \textbf{64.0\src{bench/results/v3/batch\_b\_report.json\#fresh/full context all turns/by\_category/open-domain}} & \textbf{88.2\src{bench/results/v3/batch\_b\_report.json\#fresh/full context all turns/by\_category/single-hop}} \\
\bottomrule
\end{tabular}
\end{table*}
\subsection{Robustness: a second answer model and blind human grading}\label{sec:5.4}
With Llama 3.3 70B Instruct answering from the same contexts, H1 still passes. Turns + Jev scores 75.3\%\src{bench/results/v3/second\_model\_report.json\#H1/llama-3.3-70b/acc\_a} and engram v2 75.6\%\src{bench/results/v3/second\_model\_report.json\#H1/llama-3.3-70b/acc\_b}. The difference is $-$0.3\src{bench/results/v3/second\_model\_report.json\#H1/llama-3.3-70b/d\_bar}, with one-sided bound $-$2.9\src{bench/results/v3/second\_model\_report.json\#H1/llama-3.3-70b/lower\_bound\_95\_one\_sided}. S1 also passes: Turns + Jev scores 74.9\%\src{bench/results/v3/second\_model\_report.json\#S1/llama-3.3-70b/acc\_a} and Turns + cosine 57.6\%\src{bench/results/v3/second\_model\_report.json\#S1/llama-3.3-70b/acc\_b} (p = 5.8e$-$26\src{bench/results/v3/second\_model\_report.json\#S1/llama-3.3-70b/p\_two\_sided}). Both are therefore model-robust by the registered rule. Of 3,112\src{bench/results/v3/second\_model\_report.json (four arms x 778 questions)} rebuilt contexts, all but 4\src{bench/results/v3/second\_model\_report.json\#context\_mismatches} matched their recorded token counts exactly; the four come from near-tie reorderings. OpenRouter served Llama through 9\src{bench/results/v3/second\_model\_report.json\#providers (count)} providers whose numeric precision may differ.

The first author graded, blind, both answers to each of H1's 142\src{bench/results/v3/batch\_b\_report.json\#H1 (only\_a + only\_b)} judge-discordant questions (284\src{bench/results/v3/human\_audit/audit\_report.json\#rows} rows; one question was left ungraded). Agreement with the judge was 81\%\src{bench/results/v3/human\_audit/audit\_report.json\#strict/agreement\_with\_judge} under the strict mapping and 79\%\src{bench/results/v3/human\_audit/audit\_report.json\#lenient/agreement\_with\_judge} under the lenient one (\hyperref[app:C]{Appendix~C}). Many judge-discordant pairs were not discordant to the human grader: under the strict mapping both answers were correct for 17\src{bench/results/v3/human\_audit/audit\_report.json\#strict/human\_both\_correct} questions and both wrong for 18\src{bench/results/v3/human\_audit/audit\_report.json\#strict/human\_both\_wrong}. The judge credited Turns + Jev's short answers more readily and engram v2's list-style answers less (agreement on engram v2's answers 77\%\src{bench/results/v3/human\_audit/audit\_report.json\#lenient/agreement\_with\_judge\_by\_system/engram} under the lenient mapping, against 82\%\src{bench/results/v3/human\_audit/audit\_report.json\#lenient/agreement\_with\_judge\_by\_system/T0R} on Turns + Jev's), which is why human grading widens the gap.

\subsection{Long histories (LongMemEval)}\label{sec:5.5}
LongMemEval compares Turns + Jev with mem0 and Turns + cosine only; engram v2 was not run on it, so these results cannot support any claim that Turns + Jev matches LLM-extracted memory on long histories. What they support is narrower: on histories of about 111,770\src{bench/results/v3/batch\_c\_report.json\#expansion/full context (non-abstention)/tokens\_mean} rendered tokens, the rerank still beats similarity search (S7), and no difference from mem0 was detected on knowledge-update questions (S5).

\begin{table*}[tbp]
\centering
\footnotesize\hyphenpenalty=10000\exhyphenpenalty=10000\setlength{\tabcolsep}{3.5pt}
\caption{LongMemEval, all 500\src{bench/results/v3/batch\_c\_report.json (470 scored + 30 abstention)} questions, user and assistant turns ingested: accuracy (\%) on the 470\src{bench/results/v3/batch\_c\_report.json\#S7/questions} non-abstention questions and by question type (n in parentheses), and the share of the 30\src{bench/results/v3/batch\_c\_report.json\#expansion/T0R 3 (abstention, correct = abstained)/questions} abstention questions answered by abstaining. KU: knowledge update; MS: multi-session; SS-A, SS-P, SS-U: single-session assistant, preference and user; TR: temporal reasoning; Abs: abstention. Bold: best in column within the budget group.}\label{tab:4}
\begin{tabular}{@{}>{\raggedright\arraybackslash}p{124.7pt}>{\raggedleft\arraybackslash}p{44.3pt}>{\raggedleft\arraybackslash}p{35.3pt}>{\raggedleft\arraybackslash}p{21.3pt}>{\raggedleft\arraybackslash}p{25.8pt}>{\raggedleft\arraybackslash}p{28.8pt}>{\raggedleft\arraybackslash}p{28.8pt}>{\raggedleft\arraybackslash}p{28.8pt}>{\raggedleft\arraybackslash}p{25.8pt}>{\raggedleft\arraybackslash}p{28.5pt}@{}}
\toprule
\textbf{System} & \textbf{Accuracy} & \textbf{Tokens} & \textbf{KU (72\src{bench/results/v3/batch\_c\_report.json\#expansion/L0 3 (non-abstention)/by\_type/knowledge-update/n})} & \textbf{MS (121\src{bench/results/v3/batch\_c\_report.json\#expansion/L0 3 (non-abstention)/by\_type/multi-session/n})} & \textbf{SS-A (56\src{bench/results/v3/batch\_c\_report.json\#expansion/L0 3 (non-abstention)/by\_type/single-session-assistant/n})} & \textbf{SS-P (30\src{bench/results/v3/batch\_c\_report.json\#expansion/L0 3 (non-abstention)/by\_type/single-session-preference/n})} & \textbf{SS-U (64\src{bench/results/v3/batch\_c\_report.json\#expansion/L0 3 (non-abstention)/by\_type/single-session-user/n})} & \textbf{TR (127\src{bench/results/v3/batch\_c\_report.json\#expansion/L0 3 (non-abstention)/by\_type/temporal-reasoning/n})} & \textbf{Abs} \\
\midrule
\multicolumn{10}{@{}l}{\emph{Tight budget ($k{=}$3)}} \\
Turns + cosine, $k{=}$3 & 57.7\%\src{bench/results/v3/batch\_c\_report.json\#expansion/L0 3 (non-abstention)/accuracy} & 522\src{bench/results/v3/batch\_c\_report.json\#expansion/L0 3 (non-abstention)/tokens\_mean} & 58.3\src{bench/results/v3/batch\_c\_report.json\#expansion/L0 3 (non-abstention)/by\_type/knowledge-update/accuracy} & 31.4\src{bench/results/v3/batch\_c\_report.json\#expansion/L0 3 (non-abstention)/by\_type/multi-session/accuracy} & 85.7\src{bench/results/v3/batch\_c\_report.json\#expansion/L0 3 (non-abstention)/by\_type/single-session-assistant/accuracy} & \textbf{53.3\src{bench/results/v3/batch\_c\_report.json\#expansion/L0 3 (non-abstention)/by\_type/single-session-preference/accuracy}} & 95.3\src{bench/results/v3/batch\_c\_report.json\#expansion/L0 3 (non-abstention)/by\_type/single-session-user/accuracy} & 52.0\src{bench/results/v3/batch\_c\_report.json\#expansion/L0 3 (non-abstention)/by\_type/temporal-reasoning/accuracy} & \textbf{43.3\%\src{bench/results/v3/batch\_c\_report.json\#expansion/L0 3 (abstention, correct = abstained)/accuracy}} \\
Turns + Jev, $k{=}$3 & \textbf{66.8\%\src{bench/results/v3/batch\_c\_report.json\#expansion/T0R 3 (non-abstention)/accuracy}} & 534\src{bench/results/v3/batch\_c\_report.json\#expansion/T0R 3 (non-abstention)/tokens\_mean} & \textbf{77.8\src{bench/results/v3/batch\_c\_report.json\#expansion/T0R 3 (non-abstention)/by\_type/knowledge-update/accuracy}} & \textbf{47.9\src{bench/results/v3/batch\_c\_report.json\#expansion/T0R 3 (non-abstention)/by\_type/multi-session/accuracy}} & \textbf{98.2\src{bench/results/v3/batch\_c\_report.json\#expansion/T0R 3 (non-abstention)/by\_type/single-session-assistant/accuracy}} & 46.7\src{bench/results/v3/batch\_c\_report.json\#expansion/T0R 3 (non-abstention)/by\_type/single-session-preference/accuracy} & \textbf{98.4\src{bench/results/v3/batch\_c\_report.json\#expansion/T0R 3 (non-abstention)/by\_type/single-session-user/accuracy}} & \textbf{53.5\src{bench/results/v3/batch\_c\_report.json\#expansion/T0R 3 (non-abstention)/by\_type/temporal-reasoning/accuracy}} & \textbf{43.3\%\src{bench/results/v3/batch\_c\_report.json\#expansion/T0R 3 (abstention, correct = abstained)/accuracy}} \\
\midrule
\multicolumn{10}{@{}l}{\emph{Generous budget ($k{=}$20, and full context)}} \\
Turns + cosine, $k{=}$20 & 72.8\%\src{bench/results/v3/batch\_c\_report.json\#expansion/L0 20 (non-abstention)/accuracy} & 4,352\src{bench/results/v3/batch\_c\_report.json\#expansion/L0 20 (non-abstention)/tokens\_mean} & \textbf{84.7\src{bench/results/v3/batch\_c\_report.json\#expansion/L0 20 (non-abstention)/by\_type/knowledge-update/accuracy}} & 58.7\src{bench/results/v3/batch\_c\_report.json\#expansion/L0 20 (non-abstention)/by\_type/multi-session/accuracy} & \textbf{98.2\src{bench/results/v3/batch\_c\_report.json\#expansion/L0 20 (non-abstention)/by\_type/single-session-assistant/accuracy}} & 40.0\src{bench/results/v3/batch\_c\_report.json\#expansion/L0 20 (non-abstention)/by\_type/single-session-preference/accuracy} & \textbf{96.9\src{bench/results/v3/batch\_c\_report.json\#expansion/L0 20 (non-abstention)/by\_type/single-session-user/accuracy}} & \textbf{63.8\src{bench/results/v3/batch\_c\_report.json\#expansion/L0 20 (non-abstention)/by\_type/temporal-reasoning/accuracy}} & \textbf{70.0\%\src{bench/results/v3/batch\_c\_report.json\#expansion/L0 20 (abstention, correct = abstained)/accuracy}} \\
Turns + Jev, $k{=}$20 & \textbf{73.8\%\src{bench/results/v3/batch\_c\_report.json\#expansion/T0R 20 (non-abstention)/accuracy}} & 2,278\src{bench/results/v3/batch\_c\_report.json\#expansion/T0R 20 (non-abstention)/tokens\_mean} & \textbf{84.7\src{bench/results/v3/batch\_c\_report.json\#expansion/T0R 20 (non-abstention)/by\_type/knowledge-update/accuracy}} & \textbf{67.8\src{bench/results/v3/batch\_c\_report.json\#expansion/T0R 20 (non-abstention)/by\_type/multi-session/accuracy}} & 96.4\src{bench/results/v3/batch\_c\_report.json\#expansion/T0R 20 (non-abstention)/by\_type/single-session-assistant/accuracy} & 46.7\src{bench/results/v3/batch\_c\_report.json\#expansion/T0R 20 (non-abstention)/by\_type/single-session-preference/accuracy} & \textbf{96.9\src{bench/results/v3/batch\_c\_report.json\#expansion/T0R 20 (non-abstention)/by\_type/single-session-user/accuracy}} & 58.3\src{bench/results/v3/batch\_c\_report.json\#expansion/T0R 20 (non-abstention)/by\_type/temporal-reasoning/accuracy} & 63.3\%\src{bench/results/v3/batch\_c\_report.json\#expansion/T0R 20 (abstention, correct = abstained)/accuracy} \\
Full context & 63.0\%\src{bench/results/v3/batch\_c\_report.json\#expansion/full context (non-abstention)/accuracy} & 111,770\src{bench/results/v3/batch\_c\_report.json\#expansion/full context (non-abstention)/tokens\_mean} & 81.9\src{bench/results/v3/batch\_c\_report.json\#expansion/full context (non-abstention)/by\_type/knowledge-update/accuracy} & 45.5\src{bench/results/v3/batch\_c\_report.json\#expansion/full context (non-abstention)/by\_type/multi-session/accuracy} & 91.1\src{bench/results/v3/batch\_c\_report.json\#expansion/full context (non-abstention)/by\_type/single-session-assistant/accuracy} & \textbf{56.7\src{bench/results/v3/batch\_c\_report.json\#expansion/full context (non-abstention)/by\_type/single-session-preference/accuracy}} & 92.2\src{bench/results/v3/batch\_c\_report.json\#expansion/full context (non-abstention)/by\_type/single-session-user/accuracy} & 43.3\src{bench/results/v3/batch\_c\_report.json\#expansion/full context (non-abstention)/by\_type/temporal-reasoning/accuracy} & \textbf{70.0\%\src{bench/results/v3/batch\_c\_report.json\#expansion/full context (abstention)/accuracy}} \\
\bottomrule
\end{tabular}
\end{table*}
Full context scored 63.0\%\src{bench/results/v3/batch\_c\_report.json\#expansion/full context (non-abstention)/accuracy}, against 66.8\%\src{bench/results/v3/batch\_c\_report.json\#expansion/T0R 3 (non-abstention)/accuracy} for Turns + Jev at $k{=}$3. 72\src{bench/results/v3/batch\_c\_report.json\#expansion/T0R vs full context (descriptive)/only\_a} questions were correct only for Turns + Jev and 54\src{bench/results/v3/batch\_c\_report.json\#expansion/T0R vs full context (descriptive)/only\_b} only for full context (p = 0.13\src{bench/results/v3/batch\_c\_report.json\#expansion/T0R vs full context (descriptive)/p\_two\_sided}, descriptive). Full context read 209\src{bench/results/v3/batch\_c\_report.json\#expansion (full context tokens / T0R 3 tokens)}$\times$ the tokens at 29\src{lme.cost.fc / lme.cost.t0r}$\times$ the cost per question. For reading and answering, judge excluded, it cost \$0.0169\src{bench/results/v3/full\_context\_\_lmefull\_*.json\#answers/0/query\_cost (mean, 470)} against \$0.00058\src{bench/results/v3/lean\_t0r\_\_lmefull\_*\_\_k3.json\#answers/0/query\_cost (mean, 470)}. It was weakest on temporal and multi-session questions. On the registered sample (user turns only), Turns + Jev scored 68.6\%\src{bench/results/v3/batch\_c\_report.json\#sample/T0R 3/accuracy} at $k{=}$3 and Turns + cosine 65.7\%\src{bench/results/v3/batch\_c\_report.json\#sample/L0 3/accuracy}. mem0 scored 70.0\%\src{bench/results/v3/batch\_c\_report.json\#sample/mem0 (knowledge-update 30) 3/accuracy} on the knowledge-update questions at $k{=}$3.

\subsection{Where Turns + Jev's accuracy stops}\label{sec:5.6}
Turns + Jev levels off near 77.6\%\src{bench/results/v3/batch\_a\_report.json\#fresh/T0R k20/accuracy}. At $k{=}$20 it reads only 496\src{bench/results/v3/batch\_a\_report.json\#fresh/T0R k20/tokens\_mean} tokens, because its read path keeps the shortlisted turns scored above 0.5\src{docs/V3\_PLAN.md §1, §2, §4} plus a 10\src{docs/V3\_PLAN.md §1, §2, §4}-turn cosine floor (\cref{sec:3}). The floor is part of why Turns + Jev cannot fill $k{=}$20: when few turns clear the threshold, the context stops near the floor.

Shortlist recall (exploratory as registered) locates the loss on all nine held-out conversations (1,388\src{bench/results/v3/shortlist\_recall.json\#all\_nine/all/questions} questions). All evidence turns were in the 30\src{docs/V3\_PLAN.md §1, §2, §4}-turn shortlist for 76.9\%\src{bench/results/v3/shortlist\_recall.json\#all\_nine/all/shortlist\_recall\_all} of questions. At least one was there for 88.5\%\src{bench/results/v3/shortlist\_recall.json\#all\_nine/all/shortlist\_recall\_any}. Among questions with evidence in the shortlist, the rerank kept none of it for 10.4\%\src{bench/results/v3/shortlist\_recall.json\#all\_nine/all/rerank\_drops\_all\_shortlisted\_evidence}. So about 11.5\%\src{bench/results/v3/shortlist\_recall.json\#all\_nine/all (1 - shortlist\_recall\_any)} of questions are lost to the shortlist, and another 9.2\%\src{bench/results/v3/shortlist\_recall.json\#all\_nine/all (shortlist\_recall\_any x rerank\_drops\_all\_shortlisted\_evidence)} to the rerank.

\begin{figure}[tbp]
\centering
\includegraphics[width=\columnwidth]{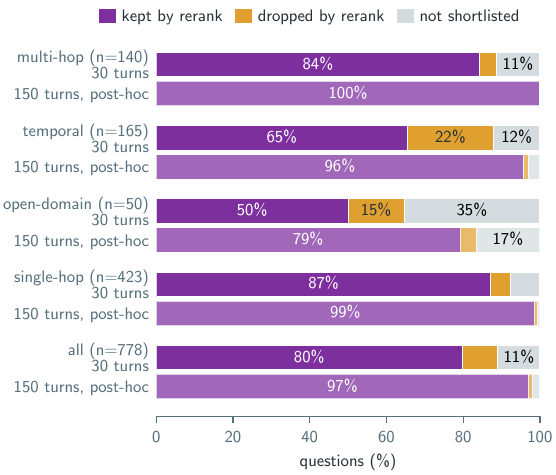}
\caption{Where Turns + Jev's accuracy stops, by category, on the 778\src{bench/results/v3/shortlist\_recall.json\#fresh\_five/all/questions} questions of the five held-out conversations: the rerank kept at least one evidence turn (purple), evidence was in the shortlist but the rerank kept none of it (amber), or no evidence turn was in the shortlist (grey). Upper bar of each pair: Turns + Jev's 30\src{docs/V3\_PLAN.md §1, §2, §4}-turn shortlist (shortlist recall, exploratory as registered). Lower, lighter bar: the wide variant's 150\src{bench/run.py\#ARMS/lean\_t0r\_wide/flags/retrieval\_shortlist}-turn shortlist (post-hoc). Percentages are printed where the segment is wide enough.}\label{fig:7}
\end{figure}

Figure~\ref{fig:7} shows the same decomposition by category on the five held-out conversations. Figure~\ref{fig:9} (\hyperref[app:J]{Appendix~J}) is an example of a shortlist miss: the evidence turn lies outside Turns + Jev's 30\src{docs/V3\_PLAN.md §1, §2, §4}-turn shortlist, while engram v2's fact extracted from it reaches the answer model.

The categories differ. Open-domain evidence reaches the shortlist least often (63.0\%\src{bench/results/v3/shortlist\_recall.json\#all\_nine/by\_category/open-domain/shortlist\_recall\_any}) and is dropped most often (31.4\%\src{bench/results/v3/shortlist\_recall.json\#all\_nine/by\_category/open-domain/rerank\_drops\_all\_shortlisted\_evidence}), consistent with Turns + Jev trailing engram v2 on open-domain questions. Temporal evidence usually reaches the shortlist but is dropped by the rerank for 22.7\%\src{bench/results/v3/shortlist\_recall.json\#all\_nine/by\_category/temporal/rerank\_drops\_all\_shortlisted\_evidence} of questions: a turn that only establishes when something happened does not look relevant to the question on its own. Multi-hop questions usually get some evidence into the shortlist (88.4\%\src{bench/results/v3/shortlist\_recall.json\#all\_nine/by\_category/multi-hop/shortlist\_recall\_any}) but rarely all of it (42.8\%\src{bench/results/v3/shortlist\_recall.json\#all\_nine/by\_category/multi-hop/shortlist\_recall\_all}).

\textbf{A wider read path (post-hoc exploratory).} After the registered results were in, we tested one variant once, outside the Holm family and in its own ledger. Turns + Jev (wide) takes a 150\src{bench/run.py\#ARMS/lean\_t0r\_wide/flags/retrieval\_shortlist}-turn cosine shortlist and asks Jev about every shortlisted turn. It keeps the top k by Jev's score, with no cut-off. Its $k{=}$47\src{bench/results/v3\_posthoc/report.json\#t0r\_wide\_k} was matched to Jev-Mem at $k{=}$40 (2,000\src{bench/results/v3\_posthoc/report.json\#comparison/T0R-wide 47 (POST-HOC EXPLORATORY (T0R-wide; docs/V3\_PLAN.md §12))/tokens\_mean} tokens against 1,987\src{bench/results/v3\_posthoc/token\_match.json\#comparator\_mean\_tokens}). It scored 81.5\%\src{bench/results/v3\_posthoc/report.json\#comparison/T0R-wide 47 (POST-HOC EXPLORATORY (T0R-wide; docs/V3\_PLAN.md §12))/accuracy}. Jev-Mem at $k{=}$40 scored 80.3\%\src{bench/results/v3/batch\_a\_report.json\#fresh/Jev-Mem k40/accuracy}, and engram v2 at $k{=}$20 scored 82.4\%\src{bench/results/v3/batch\_b\_report.json\#fresh/engram v2 k20/accuracy} with 1,238\src{bench/results/v3/batch\_b\_report.json\#fresh/engram v2 k20/tokens\_mean} tokens. All-evidence recall rose to 91.9\%\src{bench/results/v3\_posthoc/report.json\#shortlist\_recall/all/shortlist\_recall\_all}, and the rerank's losses fell to 0.9\%\src{bench/results/v3\_posthoc/report.json\#shortlist\_recall/all/top\_k\_drops\_all\_shortlisted\_evidence} (Figure~\ref{fig:7}, lower bars). This suggests the ceiling comes from Turns + Jev's read path rather than from storing raw turns. It is a hypothesis for new data, not a finding of this study.

\subsection{Cost and latency}\label{sec:5.7}
\begin{table*}[tbp]
\centering
\footnotesize\hyphenpenalty=10000\exhyphenpenalty=10000\setlength{\tabcolsep}{3.5pt}
\caption{Cost and latency (exploratory as registered), all at $k{=}$3 (the tight budget). Write cost per 1,000 turns at list prices, split by LLM, Jev and embeddings; read cost per query; read latency measured live on a fixed sample of 40\src{docs/V3\_PLAN.md §7 (fixed sample of 40 questions); bench/v3\_latency.py} questions at $k{=}$3, one query at a time, including the query-embedding call. Jev-Mem's latency comes from its reads of the same questions, measured live when they ran (42\src{bench/results/v3/read\_latency\_live.json\#Jev-Mem (3, live at run time)/queries} reads: two question texts repeat). A write-latency range spans the conversations and is compared by its lower end. $\approx$0$^\dagger$: embedding only, no model call on the read path, only an unpriced query embedding. Bold: best in column within the budget group (lowest cost and latency); read costs are not bolded, as in Table~\ref{tab:3}.}\label{tab:5}
\begin{tabular}{@{}>{\raggedright\arraybackslash}p{70.8pt}>{\raggedleft\arraybackslash}p{32.9pt}>{\raggedleft\arraybackslash}p{28.4pt}>{\raggedleft\arraybackslash}p{28.4pt}>{\raggedleft\arraybackslash}p{52.5pt}>{\raggedleft\arraybackslash}p{30.6pt}>{\raggedleft\arraybackslash}p{37.4pt}>{\raggedright\arraybackslash}p{61.7pt}>{\raggedleft\arraybackslash}p{24.6pt}>{\raggedleft\arraybackslash}p{24.6pt}@{}}
\toprule
\textbf{System} & \textbf{Write \$/1k turns} & \textbf{LLM} & \textbf{Jev} & \textbf{Embeddings} & \textbf{Write p50 (s)} & \textbf{Read \$/query} & \textbf{Jev calls/query} & \textbf{Read p50 (ms)} & \textbf{Read p90 (ms)} \\
\midrule
Turns + Jev & \textbf{\$0.0006\src{bench/results/v3/batch\_a\_report.json\#write/L0 / T0R (one store)/write\_cost\_per\_1k/embeddings}} & -- & -- & \textbf{\$0.0006\src{bench/results/v3/batch\_a\_report.json\#write/L0 / T0R (one store)/write\_cost\_per\_1k/embeddings}} & \textbf{0.2\src{bench/results/v3/batch\_a\_report.json\#write/L0 / T0R (one store)/write\_latency\_p50\_ms\_by\_conv (median, s)}} & \$0.00020\src{bench/results/v3/batch\_a\_report.json\#fresh/T0R k3/read\_cost\_per\_query} & one & 273\src{bench/results/v3/read\_latency\_live.json\#T0R/p50\_ms} & 359\src{bench/results/v3/read\_latency\_live.json\#T0R/p90\_ms} \\
Turns + cosine & \textbf{\$0.0006\src{bench/results/v3/batch\_a\_report.json\#write/L0 / T0R (one store)/write\_cost\_per\_1k/embeddings}} & -- & -- & \textbf{\$0.0006\src{bench/results/v3/batch\_a\_report.json\#write/L0 / T0R (one store)/write\_cost\_per\_1k/embeddings}} & \textbf{0.2\src{bench/results/v3/batch\_a\_report.json\#write/L0 / T0R (one store)/write\_latency\_p50\_ms\_by\_conv (median, s)}} & $\approx$0$^\dagger$ & none & \textbf{219\src{bench/results/v3/read\_latency\_live.json\#L0/p50\_ms}} & \textbf{263\src{bench/results/v3/read\_latency\_live.json\#L0/p90\_ms}} \\
Turns + LLM & \textbf{\$0.0006\src{bench/results/v3/batch\_a\_report.json\#write/L0 / T0R (one store)/write\_cost\_per\_1k/embeddings}} & -- & -- & \textbf{\$0.0006\src{bench/results/v3/batch\_a\_report.json\#write/L0 / T0R (one store)/write\_cost\_per\_1k/embeddings}} & \textbf{0.2\src{bench/results/v3/batch\_a\_report.json\#write/L0 / T0R (one store)/write\_latency\_p50\_ms\_by\_conv (median, s)}} & \$0.00023\src{bench/results/v3/batch\_b\_report.json\#fresh/T0R-LLM k3/read\_cost\_per\_query} & one (no-op) & 816\src{bench/results/v3/read\_latency\_live.json\#T0R-LLM/p50\_ms} & 1,107\src{bench/results/v3/read\_latency\_live.json\#T0R-LLM/p90\_ms} \\
engram v2 & \$1.865\src{bench/results/v3/batch\_b\_report.json\#write/engram v2/write\_cost\_per\_1k (sum)} & \$1.322\src{bench/results/v3/batch\_b\_report.json\#write/engram v2/write\_cost\_per\_1k (extraction + escalations + llm\_decisions)} & \$0.542\src{bench/results/v3/batch\_b\_report.json\#write/engram v2/write\_cost\_per\_1k/jev} & \$0.0015\src{bench/results/v3/batch\_b\_report.json\#write/engram v2/write\_cost\_per\_1k/embeddings} & 2.0\src{bench/results/v3/batch\_b\_report.json\#write/engram v2/write\_latency\_p50\_ms\_by\_conv (min, s)}--2.1\src{bench/results/v3/batch\_b\_report.json\#write/engram v2/write\_latency\_p50\_ms\_by\_conv (max, s)} & \$0.00018\src{bench/results/v3/batch\_b\_report.json\#fresh/engram v2 k3/read\_cost\_per\_query} & one & 276\src{bench/results/v3/read\_latency\_live.json\#engram v2/p50\_ms} & 319\src{bench/results/v3/read\_latency\_live.json\#engram v2/p90\_ms} \\
mem0 & \$1.321\src{bench/results/v3/batch\_b\_report.json\#write/mem0/write\_cost\_per\_1k (sum)} & \textbf{\$1.319\src{bench/results/v3/batch\_b\_report.json\#write/mem0/write\_cost\_per\_1k (extraction + escalations + llm\_decisions)}} & -- & \$0.0017\src{bench/results/v3/batch\_b\_report.json\#write/mem0/write\_cost\_per\_1k/embeddings} & 2.1\src{bench/results/v3/batch\_b\_report.json\#write/mem0/write\_latency\_p50\_ms\_by\_conv (min, s)}--2.3\src{bench/results/v3/batch\_b\_report.json\#write/mem0/write\_latency\_p50\_ms\_by\_conv (max, s)} & $\approx$0$^\dagger$ & none & 486\src{bench/results/v3/read\_latency\_live.json\#mem0/p50\_ms} & 718\src{bench/results/v3/read\_latency\_live.json\#mem0/p90\_ms} \\
Jev-Mem & \$0.212\src{bench/results/v3/batch\_a\_report.json\#write/Jev-Mem/write\_cost\_per\_1k (jev + embeddings)} & -- & \textbf{\$0.211\src{bench/results/v3/batch\_a\_report.json\#write/Jev-Mem/write\_cost\_per\_1k/jev}} & \$0.0011\src{bench/results/v3/batch\_a\_report.json\#write/Jev-Mem/write\_cost\_per\_1k/embeddings} & 0.5\src{bench/results/v3/batch\_a\_report.json\#write/Jev-Mem/write\_latency\_p50\_ms\_by\_conv (median, s)} & \$0.00120\src{bench/results/v3/batch\_a\_report.json\#fresh/Jev-Mem k3/read\_cost\_per\_query} & 5.3\src{bench/results/v3/batch\_a\_report.json\#jevmem\_reads/k3/jev\_calls\_mean} & 1,329\src{bench/results/v3/read\_latency\_live.json\#Jev-Mem (3, live at run time)/p50\_ms} & 1,622\src{bench/results/v3/read\_latency\_live.json\#Jev-Mem (3, live at run time)/p90\_ms} \\
\bottomrule
\end{tabular}
\end{table*}
Jev reads about 3.0\src{bench/results/v3/read\_latency\_live.json (T0R-LLM p50 / T0R p50)}$\times$ faster than the LLM reranker and 4.9\src{bench/results/v3/read\_latency\_live.json (Jev-Mem p50 / T0R p50)}$\times$ faster than Jev-Mem at $k{=}$3. engram v2 reads as fast as Turns + Jev, because its read path is the same one request. The write cost is where the systems differ. Extraction makes engram v2 and mem0 thousands of times more expensive to write than Turns + Jev. Jev-Mem's two Jev requests per turn cost \$0.211\src{bench/results/v3/batch\_a\_report.json\#write/Jev-Mem/write\_cost\_per\_1k/jev} per 1,000 turns. At $k{=}$40, Jev-Mem averaged 3.0\src{bench/results/v3/batch\_a\_report.json\#jevmem\_reads/k40/jev\_calls\_mean} Jev calls per query, up to 11\src{bench/results/v3/batch\_a\_report.json\#jevmem\_reads/k40/jev\_calls\_max}. Its read cost was \$0.00174\src{bench/results/v3/batch\_a\_report.json\#fresh/Jev-Mem k40/read\_cost\_per\_query} per query. Mem0's read cost is a query embedding only.

\begin{figure}[tbp]
\centering
\includegraphics[width=\columnwidth]{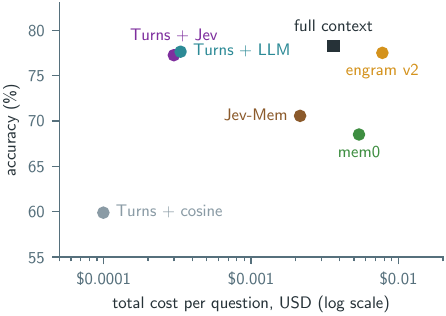}
\caption{Accuracy against total cost per question (log scale) at $k{=}$3, on the 778\src{bench/results/v3/batch\_b\_report.json\#H1/questions} questions of the five held-out conversations (exploratory as registered): write cost amortised at the benchmark's 4.0\src{bench/results/v3/batch\_b\_report.json\#write/engram v2/turns / 778} turns written per question, plus read cost and answer cost (judge excluded), at list prices; full context has no write or read cost. In a read-heavy use with one turn written per question, engram v2's total falls to \$0.00216\src{write per 1k x read\_heavy turns per question / 1000 + mean query\_cost of e4\_frozen\_sameattr\_\_k3}, mem0's to \$0.00142\src{write per 1k x read\_heavy turns per question / 1000 + mean query\_cost of mem0\_\_k3} and Jev-Mem's to \$0.00151\src{write per 1k x read\_heavy turns per question / 1000 + mean query\_cost of jevmem\_\_k3}; the other systems' totals do not change at this precision.}\label{fig:8}
\end{figure}
Figure~\ref{fig:8} plots the cost per question of \cref{eq:cost} at the benchmark's own ratio, $r_w$ = 4.0\src{bench/results/v3/batch\_b\_report.json\#write/engram v2/turns / 778}. At that ratio, Turns + Jev's total cost per question is \$0.00030\src{write per 1k x bench turns per question / 1000 + mean query\_cost of lean\_t0r\_\_k3} and engram v2's is \$0.00778\src{write per 1k x bench turns per question / 1000 + mean query\_cost of e4\_frozen\_sameattr\_\_k3}. Full context costs \$0.00362\src{write per 1k x bench turns per question / 1000 + mean query\_cost of full\_context}. In a read-heavy use, with one turn written per question, the write cost weighs less: engram v2's total falls to \$0.00216\src{write per 1k x read\_heavy turns per question / 1000 + mean query\_cost of e4\_frozen\_sameattr\_\_k3}.

\subsection{Abstention}\label{sec:5.8}
\begin{table}[tbp]
\centering
\footnotesize\hyphenpenalty=10000\exhyphenpenalty=10000\setlength{\tabcolsep}{6.0pt}
\caption{Share of LoCoMo adversarial questions (209\src{bench/results/v3/batch\_b\_report.json\#fresh\_adversarial/mem0 k3/questions}, five held-out conversations) answered by abstaining (exploratory as registered). Each column is a budget group. Bold: best in column within the budget group (highest share).}\label{tab:6}
\begin{tabular}{@{}>{\raggedright\arraybackslash}p{116.0pt}>{\raggedleft\arraybackslash}p{39.5pt}>{\raggedleft\arraybackslash}p{39.5pt}@{}}
\toprule
\textbf{System} & \textbf{$k{=}$3} & \textbf{$k{=}$20} \\
\midrule
Turns + cosine & \textbf{63.6\%\src{bench/results/v3/batch\_a\_report.json\#fresh\_adversarial/L0 k3/abstained}} & 47.8\%\src{bench/results/v3/batch\_a\_report.json\#fresh\_adversarial/L0 k20/abstained} \\
engram v2 & 59.8\%\src{bench/results/v3/batch\_b\_report.json\#fresh\_adversarial/engram v2 k3/abstained} & 52.2\%\src{bench/results/v3/batch\_b\_report.json\#fresh\_adversarial/engram v2 k20/abstained} \\
mem0 & 59.8\%\src{bench/results/v3/batch\_b\_report.json\#fresh\_adversarial/mem0 k3/abstained} & \textbf{53.6\%\src{bench/results/v3/batch\_b\_report.json\#fresh\_adversarial/mem0 k20/abstained}} \\
Turns + Jev & 54.1\%\src{bench/results/v3/batch\_a\_report.json\#fresh\_adversarial/T0R k3/abstained} & 49.3\%\src{bench/results/v3/batch\_a\_report.json\#fresh\_adversarial/T0R k20/abstained} \\
Turns + LLM & 48.8\%\src{bench/results/v3/batch\_b\_report.json\#fresh\_adversarial/T0R-LLM k3/abstained} & 53.1\%\src{bench/results/v3/batch\_b\_report.json\#fresh\_adversarial/T0R-LLM k20/abstained} \\
\bottomrule
\end{tabular}
\end{table}
Reranking lowered correct abstention at $k{=}$3. Similarity search abstained correctly on 63.6\%\src{bench/results/v3/batch\_a\_report.json\#fresh\_adversarial/L0 k3/abstained} of adversarial questions. With Jev it was 54.1\%\src{bench/results/v3/batch\_a\_report.json\#fresh\_adversarial/T0R k3/abstained}, and with an LLM reranker 48.8\%\src{bench/results/v3/batch\_b\_report.json\#fresh\_adversarial/T0R-LLM k3/abstained}. Relevant-looking context makes the answer model less willing to say that something was not mentioned. The exploratory conversations show the same (54.2\%\src{bench/results/v3/batch\_a\_report.json\#exploratory/adversarial/T0R k3/abstained} against 65.3\%\src{bench/results/v3/batch\_a\_report.json\#exploratory/adversarial/L0 k3/abstained}). So does LongMemEval at $k{=}$20: 63.3\%\src{bench/results/v3/batch\_c\_report.json\#expansion/T0R 20 (abstention, correct = abstained)/accuracy} for Turns + Jev against 70.0\%\src{bench/results/v3/batch\_c\_report.json\#expansion/L0 20 (abstention, correct = abstained)/accuracy} for Turns + cosine. Fidelity Before Structure reports that verbatim chunks abstain worse than extracted artifacts; we find that reranking adds to that.

\section{Discussion}\label{sec:6}
\textbf{A budget reading of two prior findings (interpretation).} SmartSearch finds ranking to be the bottleneck; Fidelity finds reranking marginal. Our within-study results suggest the difference is the budget. Reranking matters in proportion to how hard truncation cuts the candidate set. In SmartSearch a question has about 431\src{cite:derehag2026smartsearch (checked against the paper's text; paper\_v3/bib\_verification.md)} grep candidates on average. About 62\src{cite:derehag2026smartsearch (checked against the paper's text; paper\_v3/bib\_verification.md)} passages fit its 2,000\src{cite:derehag2026smartsearch (checked against the paper's text; paper\_v3/bib\_verification.md)}-word budget. Without ranking, only 22.5\src{cite:derehag2026smartsearch (checked against the paper's text; paper\_v3/bib\_verification.md)}\% of gold evidence survives truncation. Our $k{=}$3 likewise keeps three of 30\src{docs/V3\_PLAN.md §1, §2, §4}, and both show large ranking gains. Fidelity reranks a top-30\src{cite:an2026fidelity (checked against the paper's text; paper\_v3/bib\_verification.md)} pool to 15\src{cite:an2026fidelity (checked against the paper's text; paper\_v3/bib\_verification.md)} with bge-reranker-v2-m3, under a 5,000\src{cite:an2026fidelity (checked against the paper's text; paper\_v3/bib\_verification.md)}-token cap. Its gains are 2.9\src{cite:an2026fidelity (checked against the paper's text; paper\_v3/bib\_verification.md)} points on LoCoMo and 0.6\src{cite:an2026fidelity (checked against the paper's text; paper\_v3/bib\_verification.md)} on LongMemEval-S. Our $k{=}$20 likewise keeps twenty of 30\src{docs/V3\_PLAN.md §1, §2, §4}, and both show small gains. This is an interpretation across pipelines that differ in retrievers, rerankers, answer models and judges. Inside our study it is supported by the $k{=}$3 to $k{=}$20 comparison on two benchmarks; it is not a tested claim across papers. Read together with \citet{kang2026retain}, the two studies suggest that compression is favoured when the budget cannot fit the relevant raw evidence, a regime our budgets did not reach, and that raw evidence with good selection is competitive once it can, as at our tight budgets.

\textbf{When extraction is worth it.} At the tight budget of H1, extraction adds little and costs thousands of times more to write. At generous budgets it is more accurate and more compact: engram v2 at $k{=}$20 was the most accurate system we measured, with fewer tokens than Jev-Mem at $k{=}$40. Open-domain and temporal questions are where Turns + Jev trailed engram v2, and where its shortlist and rerank lose the most evidence.

\textbf{What a typed decision model contributes.} In this study, it contributed speed and cost at equal selection quality, not higher accuracy. At matched context, Jev selected as accurately as the gpt-4o-mini reranker (S4) at about a third of the latency, in one request per question. One Jev request also beat Jev-Mem's multi-request graph walk at matched context (S2). A typed question returns a probability over fixed options in one short call, which is what a reranker needs.

\textbf{Judge leniency and answer style.} mem0's LoCoMo judge is lenient, and in our audit it credited short answers more readily than list-style ones, so its agreement with human grading differed by system. Fidelity Before Structure's human study found no such dependence on answer length, but its LoCoMo judge is instructed to be strict (``Binary - strict'': ``partial answers or answers with significant missing information should be marked INCORRECT'', their Appendix J.4), while mem0's asks the judge to ``be generous with your grading - as long as it touches on the same topic as the gold answer''. A strict instruction leaves less room for style to matter, which may explain why their audit found no short-answer bias and ours did. Other audits point the same way. On multimodal memory questions, MemLens \citep{ren2026memlens} finds that its LLM judge's leniency inflates closed-form accuracy by about 5\src{cite:ren2026memlens (checked against the paper's text; paper\_v3/bib\_verification.md)} points, without reordering its leaderboard. An audit of LoCoMo by Penfield Labs \citep{penfield2026locomo} reports 99\src{cite:penfield2026locomo (checked against the paper's text; paper\_v3/bib\_verification.md)} answer-key errors in 1,540\src{cite:penfield2026locomo (checked against the paper's text; paper\_v3/bib\_verification.md)} questions (6.4\src{cite:penfield2026locomo (checked against the paper's text; paper\_v3/bib\_verification.md)}\%) and a gpt-4o-mini judge that accepted 62.81\src{cite:penfield2026locomo (checked against the paper's text; paper\_v3/bib\_verification.md)}\% of deliberately wrong but topically adjacent answers. Memory benchmarks that compare systems with different answer styles should report judge--human agreement by system.

\textbf{Not state of the art.} SmartSearch reports 91.9\src{cite:derehag2026smartsearch (checked against the paper's text; paper\_v3/bib\_verification.md)}\% on LoCoMo under its own protocol. That protocol uses gpt-4o-mini to answer and judge, binary judgments, all ten conversations and 1,540\src{cite:derehag2026smartsearch (checked against the paper's text; paper\_v3/bib\_verification.md)} questions in categories 1--4, at 3,141\src{cite:derehag2026smartsearch (checked against the paper's text; paper\_v3/bib\_verification.md)} tokens per question. Our numbers come from a different protocol on five held-out conversations and are not comparable to it. Our best result, the post-hoc Turns + Jev (wide), is below that figure.

\section*{Limitations}
\begin{itemize}
\item \textbf{Benchmarks.} One benchmark family per setting: LoCoMo, whose dialogues are LLM-generated, and LongMemEval. Five primary conversations give 778\src{bench/results/v3/batch\_b\_report.json\#H1/questions} questions; categories are small.
\item \textbf{LongMemEval scope.} engram v2 was not run on LongMemEval, so no claim about extraction on long histories follows from this study.
\item \textbf{Human grading.} The grader, the first author, built the systems evaluated; the mapping of partial grades was not pre-specified (two are reported); one question was ungraded; and only judge-discordant questions were re-graded, so judge errors on questions where the judge agreed across systems remain.
\item \textbf{Adversarial content.} Turns + Jev passes raw, user-written turns to Jev's relevance question, so text injected into a conversation could shift which turns are selected; prompt injection shifts Jev's decision probabilities \citep{wu2026hijacking}. We did not test adversarial content.
\item \textbf{A closed decision model.} Jev is a closed, versioned model; results hold for jev-1.13.0.
\item \textbf{mem0 serving.} mem0's extraction calls were served through OpenRouter, about half by Azure, not the OpenAI API as registered; only S3 involves mem0.
\item \textbf{Post-hoc variant.} Turns + Jev (wide) was designed after the registered results and tested once on the same questions.
\item \textbf{Budget and read path.} The $k{=}$20 comparisons mix the budget with Turns + Jev's read-path ceiling: its threshold and cosine floor keep it at 496\src{bench/results/v3/batch\_a\_report.json\#fresh/T0R k20/tokens\_mean} tokens at $k{=}$20, against Turns + cosine's 826\src{bench/results/v3/batch\_a\_report.json\#fresh/L0 k20/tokens\_mean}. The budget dependence at generous budgets may be less steep for a wider read path; Turns + Jev (wide) suggests so, post-hoc.
\item \textbf{Absolute accuracy.} Below SmartSearch's reported figures at generous budgets, under a different protocol.
\item \textbf{Development data.} Every design choice was made on one conversation, conv-26.
\end{itemize}
\section{Conclusion}\label{sec:7}
Within this study, reranking's gain over similarity search shrinks as the context budget grows. Reranking added 17.4\src{bench/results/v3/batch\_a\_report.json\#fresh (T0R k3 - L0 k3)} points on LoCoMo and 9.1\src{bench/results/v3/batch\_c\_report.json\#expansion (T0R 3 - L0 3)} on LongMemEval when three of 30\src{docs/V3\_PLAN.md §1, §2, §4} candidates were kept. At $k{=}$20 it added 1.5\src{bench/results/v3/batch\_a\_report.json\#fresh (T0R k20 - L0 k20)} and 1.1\src{bench/results/v3/batch\_c\_report.json\#expansion (T0R 20 - L0 20)}, and extraction systems were more accurate. This suggests an explanation for the published disagreement, which remains an interpretation across papers; \citet{kang2026retain} find a complementary budget dependence for consolidation. A pre-registered non-inferiority test on held-out conversations bounds what extraction adds at a tight budget to at most 4.7\src{bench/results/v3/human\_audit/audit\_report.json\#lenient/H1\_with\_human\_grades/lower\_bound\_95\_one\_sided (negated)} points. That test holds with a second answer model and under blind human grading, at 3,061\src{bench/results/v3/batch\_b\_report.json\#write\_cost\_ratio\_engram\_over\_t0r/ratio}$\times$ lower write cost. A typed decision model is an effective selector. At matched context, Jev was non-inferior to an LLM reranker (bound $-$2.0\src{bench/results/v3/batch\_b\_report.json\#S4/lower\_bound\_95\_one\_sided}) at about a third of the latency. It was also more accurate than a multi-call Jev graph traversal at matched context. The diagnostics locate where selection stops: in shortlist misses (11.5\%\src{bench/results/v3/shortlist\_recall.json\#all\_nine/all (1 - shortlist\_recall\_any)}) and rerank drops (9.2\%\src{bench/results/v3/shortlist\_recall.json\#all\_nine/all (shortlist\_recall\_any x rerank\_drops\_all\_shortlisted\_evidence)}), most for temporal evidence. They also show that an LLM judge's leniency, documented elsewhere, interacts with answer length. Memory benchmarks that compare systems with different answer styles should therefore report judge--human agreement by system.

\section*{Author Contributions}
Rishabh Sharma designed the study and its pre-registered plan, built the systems, ran and orchestrated the experiments, and did the blind human audit; the human grader is therefore the author of the systems evaluated. Rishika Lall contributed to the analysis and interpretation of the results. Both authors drafted, reviewed and edited the paper.

\section*{AI Assistance}
The code, run orchestration and drafting of this paper were done with Claude Code (Anthropic) under the first author's direction. The first author made every methodological decision, approved each stage of the registered plan and did the human audit.

\section*{Artifacts}
Code, plans, per-question answers and judge labels, and the human-audit grades with their key are at github.com/ris3abh/Engram: tags \texttt{v3-frozen}, \texttt{v3-amended} and the paper tag; results in \texttt{bench/\allowbreak{}results/\allowbreak{}v3/\allowbreak{}} (per-question files, reports, ledgers, \texttt{human\_\allowbreak{}audit/\allowbreak{}}) and \texttt{bench/\allowbreak{}results/\allowbreak{}v3\_\allowbreak{}posthoc/\allowbreak{}}. The plan and its amendment are deposited at 10.5281/zenodo.22970745 and 10.5281/zenodo.22977848; this paper is 10.5281/zenodo.22985242 (release tag \texttt{paper-v3-preprint-r3}); the earlier engram preprint is 10.5281/zenodo.22941757 \citep{sharma2026typed}.

\bibliography{references}
\appendix
\onecolumn
\FloatBarrier
\section{Per-conversation results}\label{app:A}
\begin{table}[!htbp]
\centering
\footnotesize\hyphenpenalty=10000\exhyphenpenalty=10000\setlength{\tabcolsep}{6.0pt}
\caption{Accuracy (\%) per conversation, LoCoMo scored categories.}
\begin{tabular}{@{}>{\raggedright\arraybackslash}p{169.4pt}>{\raggedleft\arraybackslash}p{45.1pt}>{\raggedleft\arraybackslash}p{45.1pt}>{\raggedleft\arraybackslash}p{45.1pt}>{\raggedleft\arraybackslash}p{45.1pt}>{\raggedleft\arraybackslash}p{45.1pt}@{}}
\toprule
\textbf{System, setting} & \textbf{conv-44 (123\src{bench/results/v3/lean\_t0r\_\_heldout\_conv-44\_\_k3.json\#answers (count)})} & \textbf{conv-47 (150\src{bench/results/v3/lean\_t0r\_\_heldout\_conv-47\_\_k3.json\#answers (count)})} & \textbf{conv-48 (191\src{bench/results/v3/lean\_t0r\_\_heldout\_conv-48\_\_k3.json\#answers (count)})} & \textbf{conv-49 (156\src{bench/results/v3/lean\_t0r\_\_heldout\_conv-49\_\_k3.json\#answers (count)})} & \textbf{conv-50 (158\src{bench/results/v3/lean\_t0r\_\_heldout\_conv-50\_\_k3.json\#answers (count)})} \\
\midrule
Turns + Jev, $k{=}$3 & 78.0\src{bench/results/v3/lean\_t0r\_\_heldout\_conv-44\_\_k3.json\#answers (share CORRECT)} & 76.7\src{bench/results/v3/lean\_t0r\_\_heldout\_conv-47\_\_k3.json\#answers (share CORRECT)} & 79.1\src{bench/results/v3/lean\_t0r\_\_heldout\_conv-48\_\_k3.json\#answers (share CORRECT)} & 73.7\src{bench/results/v3/lean\_t0r\_\_heldout\_conv-49\_\_k3.json\#answers (share CORRECT)} & 78.5\src{bench/results/v3/lean\_t0r\_\_heldout\_conv-50\_\_k3.json\#answers (share CORRECT)} \\
Turns + Jev, $k{=}$6 & 79.7\src{bench/results/v3/lean\_t0r\_\_heldout\_conv-44\_\_k6.json\#answers (share CORRECT)} & 76.7\src{bench/results/v3/lean\_t0r\_\_heldout\_conv-47\_\_k6.json\#answers (share CORRECT)} & 78.5\src{bench/results/v3/lean\_t0r\_\_heldout\_conv-48\_\_k6.json\#answers (share CORRECT)} & 73.7\src{bench/results/v3/lean\_t0r\_\_heldout\_conv-49\_\_k6.json\#answers (share CORRECT)} & 76.6\src{bench/results/v3/lean\_t0r\_\_heldout\_conv-50\_\_k6.json\#answers (share CORRECT)} \\
Turns + Jev, $k{=}$20 & 79.7\src{bench/results/v3/lean\_t0r\_\_heldout\_conv-44\_\_k20.json\#answers (share CORRECT)} & 74.7\src{bench/results/v3/lean\_t0r\_\_heldout\_conv-47\_\_k20.json\#answers (share CORRECT)} & 81.7\src{bench/results/v3/lean\_t0r\_\_heldout\_conv-48\_\_k20.json\#answers (share CORRECT)} & 75.0\src{bench/results/v3/lean\_t0r\_\_heldout\_conv-49\_\_k20.json\#answers (share CORRECT)} & 76.6\src{bench/results/v3/lean\_t0r\_\_heldout\_conv-50\_\_k20.json\#answers (share CORRECT)} \\
Turns + cosine, $k{=}$3 & 65.9\src{bench/results/v3/lean\_l0\_\_heldout\_conv-44\_\_k3.json\#answers (share CORRECT)} & 57.3\src{bench/results/v3/lean\_l0\_\_heldout\_conv-47\_\_k3.json\#answers (share CORRECT)} & 62.8\src{bench/results/v3/lean\_l0\_\_heldout\_conv-48\_\_k3.json\#answers (share CORRECT)} & 58.3\src{bench/results/v3/lean\_l0\_\_heldout\_conv-49\_\_k3.json\#answers (share CORRECT)} & 55.7\src{bench/results/v3/lean\_l0\_\_heldout\_conv-50\_\_k3.json\#answers (share CORRECT)} \\
Turns + cosine, $k{=}$20 & 78.0\src{bench/results/v3/lean\_l0\_\_heldout\_conv-44\_\_k20.json\#answers (share CORRECT)} & 74.0\src{bench/results/v3/lean\_l0\_\_heldout\_conv-47\_\_k20.json\#answers (share CORRECT)} & 79.1\src{bench/results/v3/lean\_l0\_\_heldout\_conv-48\_\_k20.json\#answers (share CORRECT)} & 74.4\src{bench/results/v3/lean\_l0\_\_heldout\_conv-49\_\_k20.json\#answers (share CORRECT)} & 74.7\src{bench/results/v3/lean\_l0\_\_heldout\_conv-50\_\_k20.json\#answers (share CORRECT)} \\
Turns + LLM, $k{=}$3 & 76.4\src{bench/results/v3/lean\_t0r\_llm\_\_heldout\_conv-44\_\_k3.json\#answers (share CORRECT)} & 72.7\src{bench/results/v3/lean\_t0r\_llm\_\_heldout\_conv-47\_\_k3.json\#answers (share CORRECT)} & 80.6\src{bench/results/v3/lean\_t0r\_llm\_\_heldout\_conv-48\_\_k3.json\#answers (share CORRECT)} & 77.6\src{bench/results/v3/lean\_t0r\_llm\_\_heldout\_conv-49\_\_k3.json\#answers (share CORRECT)} & 79.7\src{bench/results/v3/lean\_t0r\_llm\_\_heldout\_conv-50\_\_k3.json\#answers (share CORRECT)} \\
engram v2, $k{=}$3 & 77.2\src{bench/results/v3/e4\_frozen\_sameattr\_\_heldout\_conv-44\_\_k3.json\#answers (share CORRECT)} & 80.7\src{bench/results/v3/e4\_frozen\_sameattr\_\_heldout\_conv-47\_\_k3.json\#answers (share CORRECT)} & 78.5\src{bench/results/v3/e4\_frozen\_sameattr\_\_heldout\_conv-48\_\_k3.json\#answers (share CORRECT)} & 76.3\src{bench/results/v3/e4\_frozen\_sameattr\_\_heldout\_conv-49\_\_k3.json\#answers (share CORRECT)} & 74.7\src{bench/results/v3/e4\_frozen\_sameattr\_\_heldout\_conv-50\_\_k3.json\#answers (share CORRECT)} \\
engram v2, $k{=}$20 & 84.6\src{bench/results/v3/e4\_frozen\_sameattr\_\_heldout\_conv-44\_\_k20.json\#answers (share CORRECT)} & 84.7\src{bench/results/v3/e4\_frozen\_sameattr\_\_heldout\_conv-47\_\_k20.json\#answers (share CORRECT)} & 82.2\src{bench/results/v3/e4\_frozen\_sameattr\_\_heldout\_conv-48\_\_k20.json\#answers (share CORRECT)} & 80.8\src{bench/results/v3/e4\_frozen\_sameattr\_\_heldout\_conv-49\_\_k20.json\#answers (share CORRECT)} & 80.4\src{bench/results/v3/e4\_frozen\_sameattr\_\_heldout\_conv-50\_\_k20.json\#answers (share CORRECT)} \\
mem0, $k{=}$3 & 68.3\src{bench/results/v3/mem0\_\_heldout\_conv-44\_\_k3.json\#answers (share CORRECT)} & 65.3\src{bench/results/v3/mem0\_\_heldout\_conv-47\_\_k3.json\#answers (share CORRECT)} & 69.6\src{bench/results/v3/mem0\_\_heldout\_conv-48\_\_k3.json\#answers (share CORRECT)} & 72.4\src{bench/results/v3/mem0\_\_heldout\_conv-49\_\_k3.json\#answers (share CORRECT)} & 66.5\src{bench/results/v3/mem0\_\_heldout\_conv-50\_\_k3.json\#answers (share CORRECT)} \\
mem0, $k{=}$20 & 79.7\src{bench/results/v3/mem0\_\_heldout\_conv-44\_\_k20.json\#answers (share CORRECT)} & 78.0\src{bench/results/v3/mem0\_\_heldout\_conv-47\_\_k20.json\#answers (share CORRECT)} & 81.2\src{bench/results/v3/mem0\_\_heldout\_conv-48\_\_k20.json\#answers (share CORRECT)} & 78.8\src{bench/results/v3/mem0\_\_heldout\_conv-49\_\_k20.json\#answers (share CORRECT)} & 75.3\src{bench/results/v3/mem0\_\_heldout\_conv-50\_\_k20.json\#answers (share CORRECT)} \\
Jev-Mem, $k{=}$3 & 69.1\src{bench/results/v3/jevmem\_\_heldout\_conv-44\_\_k3.json\#answers (share CORRECT)} & 66.7\src{bench/results/v3/jevmem\_\_heldout\_conv-47\_\_k3.json\#answers (share CORRECT)} & 74.3\src{bench/results/v3/jevmem\_\_heldout\_conv-48\_\_k3.json\#answers (share CORRECT)} & 69.2\src{bench/results/v3/jevmem\_\_heldout\_conv-49\_\_k3.json\#answers (share CORRECT)} & 72.2\src{bench/results/v3/jevmem\_\_heldout\_conv-50\_\_k3.json\#answers (share CORRECT)} \\
Jev-Mem, $k{=}$40 & 81.3\src{bench/results/v3/jevmem\_\_heldout\_conv-44\_\_k40.json\#answers (share CORRECT)} & 78.7\src{bench/results/v3/jevmem\_\_heldout\_conv-47\_\_k40.json\#answers (share CORRECT)} & 80.6\src{bench/results/v3/jevmem\_\_heldout\_conv-48\_\_k40.json\#answers (share CORRECT)} & 82.7\src{bench/results/v3/jevmem\_\_heldout\_conv-49\_\_k40.json\#answers (share CORRECT)} & 78.5\src{bench/results/v3/jevmem\_\_heldout\_conv-50\_\_k40.json\#answers (share CORRECT)} \\
Full context & 82.1\src{bench/results/v3/full\_context\_\_heldout\_conv-44.json\#answers (share CORRECT)} & 72.7\src{bench/results/v3/full\_context\_\_heldout\_conv-47.json\#answers (share CORRECT)} & 82.7\src{bench/results/v3/full\_context\_\_heldout\_conv-48.json\#answers (share CORRECT)} & 76.3\src{bench/results/v3/full\_context\_\_heldout\_conv-49.json\#answers (share CORRECT)} & 77.2\src{bench/results/v3/full\_context\_\_heldout\_conv-50.json\#answers (share CORRECT)} \\
\bottomrule
\end{tabular}
\end{table}
The exploratory replication on conv-30, conv-41, conv-42 and conv-43 (610\src{bench/results/v3/batch\_a\_report.json\#exploratory/L0 k3/questions} scored questions): Turns + cosine 60.7\%\src{bench/results/v3/batch\_a\_report.json\#exploratory/L0 k3/accuracy} at $k{=}$3 and 76.2\%\src{bench/results/v3/batch\_a\_report.json\#exploratory/L0 k20/accuracy} at $k{=}$20; Turns + Jev 77.2\%\src{bench/results/v3/batch\_a\_report.json\#exploratory/T0R k3/accuracy} at $k{=}$3 and 76.6\%\src{bench/results/v3/batch\_a\_report.json\#exploratory/T0R k20/accuracy} at $k{=}$20. Turns + Jev's matched k against Turns + cosine was 3\src{bench/results/v3/batch\_a\_report.json\#token\_match/L0 (exploratory)/t0r\_k}, with 116\src{bench/results/v3/batch\_a\_report.json\#exploratory/T0R vs L0 (matched)/only\_a} questions correct only for Turns + Jev and 15\src{bench/results/v3/batch\_a\_report.json\#exploratory/T0R vs L0 (matched)/only\_b} only for Turns + cosine (p = 1.6e$-$20\src{bench/results/v3/batch\_a\_report.json\#exploratory/T0R vs L0 (matched)/p\_two\_sided}, exploratory).

\FloatBarrier
\section{Registered plan and deviations}\label{app:B}
The plan's guarded sections in \texttt{docs/\allowbreak{}V3\_\allowbreak{}PLAN.md}, checked by a test that fails on any undated change, fixed the systems, data, token-matching rule, tests, predictions, human check, run order and budget before any run. Every later change is a dated entry in its Deviations section, one row each below; presentation changes share a row.

\begingroup\scriptsize
\begin{longtable}{>{\raggedright\arraybackslash}p{0.108\linewidth}>{\raggedright\arraybackslash}p{0.259\linewidth}>{\raggedright\arraybackslash}p{0.259\linewidth}>{\raggedright\arraybackslash}p{0.259\linewidth}}
\caption{Deviations from the registered plan.}\label{tab:8}\\
\toprule
\textbf{Date} & \textbf{Change} & \textbf{Reason} & \textbf{Effect on results} \\
\midrule
\endfirsthead
\toprule
\textbf{Date} & \textbf{Change} & \textbf{Reason} & \textbf{Effect on results} \\
\midrule
\endhead
2026-09-26 & Amendment, deposited after Batch A (S1 and S2 known) and before any H1 result: shortlist recall; LongMemEval on all 500\src{bench/results/v3/batch\_\allowbreak{}c\_\allowbreak{}report.json (470 scored + 30 abstention)} questions with user and assistant turns, with the new test S7; the second answer model; the outcome paragraphs and the rule for ``LongMemEval holds''; budget caps & Extend the study before the primary test was run & S7 joins the Holm family; new robustness checks; H1, its margin and the other tests unchanged \\
2026-09-26 & Outcome paragraphs revised before upload & The first author's own wording & None; made before any H1 result \\
2026-09-26 & mem0's extraction calls went through OpenRouter, about half served by Azure, instead of the OpenAI API; the ledger was corrected and guards added before any later run & A mem0 library default routes calls to OpenRouter when its key is set (\hyperref[app:G]{Appendix~G}) & S3 is reported with a caveat; no other test involves mem0 \\
2026-09-26 & Runs repeated after OpenAI rate limits, with more retries and a Jev throttle; completed calls replayed from the call cache & Rate limits & None: replayed calls are identical \\
2026-09-26 & Turns + LLM also asks Jev's query-relation question once per query & Shared read-path code & None: the question is a no-op on turns \\
2026-09-26 & Token counting treats text that spells a special token (\texttt{\textless{}|endoftext|\textgreater{}}, in one LongMemEval haystack) as ordinary text & The tokenizer refused that text & One question re-run; every other count unchanged \\
2026-09-27 & Human-check grades: the sheet asked for CORRECT or WRONG, and partial grades appeared, so two mappings are reported & The plan fixed no rule for partial grades & Both mappings reported (Table~\ref{tab:1}, \hyperref[app:C]{Appendix~C}); H1 is decided by the judge \\
2026-09-26 and 2026-09-27 & Presentation only: the paper title replaced; systems renamed Turns + Jev, Turns + cosine, Turns + LLM and Turns + Jev (wide) (registered as T0R, L0, T0R-LLM and T0R-wide); the audit sheet put each answer on its own row and shuffled all 284\src{bench/results/v3/human\_\allowbreak{}audit/audit\_\allowbreak{}report.json\#rows} rows, rather than shuffling within each question & The selected title presented a published idea as new and its ``matches'' overstated a non-inferiority result; readability; sheet layout & None \\
\bottomrule
\end{longtable}
\endgroup
An implementation note not in the Deviations section: two retrieval-only sweeps were stopped by mistake and re-run from the cache.

\FloatBarrier
\section{Human audit}\label{app:C}
\textbf{Protocol.} The sheet held every question on which the judge found exactly one of H1's two answers correct (142\src{bench/results/v3/batch\_b\_report.json\#H1 (only\_a + only\_b)} questions), each answer as its own row (284\src{bench/results/v3/human\_audit/audit\_report.json\#rows} rows), shuffled with seed 0, with the question and gold answer shown and no system name or judge label. The sheet asked for CORRECT or WRONG; the plan's notes had listed CORRECT, WRONG or UNCLEAR. The first author's grades included partial and hedged labels, and one question was left ungraded. Two mappings are reported: strict (only grades starting with CORRECT count as correct) and lenient (partial and hedged-correct grades also count); any grade containing WRONG counts as wrong under both. H1 is decided by the judge.

\begin{table}[!htbp]
\centering
\footnotesize\hyphenpenalty=10000\exhyphenpenalty=10000\setlength{\tabcolsep}{4.5pt}

\begin{tabular}{@{}>{\raggedright\arraybackslash}p{109.2pt}>{\raggedleft\arraybackslash}p{53.8pt}>{\raggedleft\arraybackslash}p{44.6pt}>{\raggedleft\arraybackslash}p{44.6pt}>{\raggedleft\arraybackslash}p{34.9pt}>{\raggedleft\arraybackslash}p{40.1pt}>{\raggedleft\arraybackslash}p{29.3pt}>{\raggedleft\arraybackslash}p{35.6pt}@{}}
\toprule
\textbf{Mapping} & \textbf{Agreement with judge} & \textbf{On Turns + Jev's answers} & \textbf{On engram v2's answers} & \textbf{Only Turns + Jev right} & \textbf{Only engram v2 right} & \textbf{Both right} & \textbf{Both wrong} \\
\midrule
Strict & 81\%\src{bench/results/v3/human\_audit/audit\_report.json\#strict/agreement\_with\_judge} & 81\%\src{bench/results/v3/human\_audit/audit\_report.json\#strict/agreement\_with\_judge\_by\_system/T0R} & 82\%\src{bench/results/v3/human\_audit/audit\_report.json\#strict/agreement\_with\_judge\_by\_system/engram} & 47\src{bench/results/v3/human\_audit/audit\_report.json\#strict/human\_only\_t0r} & 59\src{bench/results/v3/human\_audit/audit\_report.json\#strict/human\_only\_engram} & 17\src{bench/results/v3/human\_audit/audit\_report.json\#strict/human\_both\_correct} & 18\src{bench/results/v3/human\_audit/audit\_report.json\#strict/human\_both\_wrong} \\
Lenient & 79\%\src{bench/results/v3/human\_audit/audit\_report.json\#lenient/agreement\_with\_judge} & 82\%\src{bench/results/v3/human\_audit/audit\_report.json\#lenient/agreement\_with\_judge\_by\_system/T0R} & 77\%\src{bench/results/v3/human\_audit/audit\_report.json\#lenient/agreement\_with\_judge\_by\_system/engram} & 41\src{bench/results/v3/human\_audit/audit\_report.json\#lenient/human\_only\_t0r} & 60\src{bench/results/v3/human\_audit/audit\_report.json\#lenient/human\_only\_engram} & 31\src{bench/results/v3/human\_audit/audit\_report.json\#lenient/human\_both\_correct} & 9\src{bench/results/v3/human\_audit/audit\_report.json\#lenient/human\_both\_wrong} \\
\bottomrule
\end{tabular}
\end{table}
\textbf{The ungraded question.} One discordant question, in conv-47, has an ungraded row. H1 with human grades can treat it two ways: (a) it keeps the judge's labels, or (b) it is dropped. The paper reports (a) in Table~\ref{tab:1} and \cref{sec:5.4}. (a) keeps all 778\src{bench/results/v3/batch\_b\_report.json\#H1/questions} questions, and it is the conservative choice: the judge scored only engram v2 correct on this question. The strict 78.0\%\src{bench/results/v3/human\_audit/audit\_report.json\#strict/H1\_with\_human\_grades/acc\_engram} and lenient 79.9\%\src{bench/results/v3/human\_audit/audit\_report.json\#lenient/H1\_with\_human\_grades/acc\_engram} for engram v2 are the (a) values.

\begin{table}[!htbp]
\centering
\footnotesize\hyphenpenalty=10000\exhyphenpenalty=10000\setlength{\tabcolsep}{4.5pt}
\caption{H1 with human grades under both treatments of the ungraded question. Differences and bounds in points.}\label{tab:9}
\begin{tabular}{@{}>{\raggedright\arraybackslash}p{134.7pt}>{\raggedleft\arraybackslash}p{45.9pt}>{\raggedleft\arraybackslash}p{31.5pt}>{\raggedleft\arraybackslash}p{36.7pt}>{\raggedleft\arraybackslash}p{52.2pt}>{\raggedleft\arraybackslash}p{46.2pt}>{\raggedleft\arraybackslash}p{53.7pt}@{}}
\toprule
\textbf{Mapping, treatment} & \textbf{Questions} & \textbf{Turns + Jev} & \textbf{engram v2} & \textbf{Difference} & \textbf{One-sided 95\% bound} & \textbf{Two-sided 95\% CI} \\
\midrule
Strict, (a) judge's labels & 778\src{bench/results/v3/batch\_b\_report.json\#H1/questions} & 76.3\%\src{bench/results/v3/human\_audit/audit\_report.json\#strict/H1\_with\_human\_grades/acc\_t0r} & 78.0\%\src{bench/results/v3/human\_audit/audit\_report.json\#strict/H1\_with\_human\_grades/acc\_engram} & $-$1.7\src{bench/results/v3/human\_audit/audit\_report.json\#strict/H1\_with\_human\_grades/d\_bar} & $-$3.9\src{bench/results/v3/human\_audit/audit\_report.json\#strict/H1\_with\_human\_grades/lower\_bound\_95\_one\_sided} & [$-$4.3\src{bench/results/v3/human\_audit/audit\_report.json\#strict/H1\_with\_human\_grades/ci95\_two\_sided/0},~+0.9\src{bench/results/v3/human\_audit/audit\_report.json\#strict/H1\_with\_human\_grades/ci95\_two\_sided/1}] \\
Strict, (b) dropped & 777\src{bench/results/v3/human\_audit/audit\_report.json\#strict/H1\_with\_human\_grades\_ungraded\_dropped/questions} & 76.4\%\src{bench/results/v3/human\_audit/audit\_report.json\#strict/H1\_with\_human\_grades\_ungraded\_dropped/acc\_t0r} & 78.0\%\src{bench/results/v3/human\_audit/audit\_report.json\#strict/H1\_with\_human\_grades\_ungraded\_dropped/acc\_engram} & $-$1.5\src{bench/results/v3/human\_audit/audit\_report.json\#strict/H1\_with\_human\_grades\_ungraded\_dropped/d\_bar} & $-$3.7\src{bench/results/v3/human\_audit/audit\_report.json\#strict/H1\_with\_human\_grades\_ungraded\_dropped/lower\_bound\_95\_one\_sided} & [$-$4.1\src{bench/results/v3/human\_audit/audit\_report.json\#strict/H1\_with\_human\_grades\_ungraded\_dropped/ci95\_two\_sided/0},~+1.1\src{bench/results/v3/human\_audit/audit\_report.json\#strict/H1\_with\_human\_grades\_ungraded\_dropped/ci95\_two\_sided/1}] \\
Lenient, (a) judge's labels & 778\src{bench/results/v3/batch\_b\_report.json\#H1/questions} & 77.4\%\src{bench/results/v3/human\_audit/audit\_report.json\#lenient/H1\_with\_human\_grades/acc\_t0r} & 79.9\%\src{bench/results/v3/human\_audit/audit\_report.json\#lenient/H1\_with\_human\_grades/acc\_engram} & $-$2.6\src{bench/results/v3/human\_audit/audit\_report.json\#lenient/H1\_with\_human\_grades/d\_bar} & $-$4.7\src{bench/results/v3/human\_audit/audit\_report.json\#lenient/H1\_with\_human\_grades/lower\_bound\_95\_one\_sided} & [$-$5.1\src{bench/results/v3/human\_audit/audit\_report.json\#lenient/H1\_with\_human\_grades/ci95\_two\_sided/0},~$-$0.03\src{bench/results/v3/human\_audit/audit\_report.json\#lenient/H1\_with\_human\_grades/ci95\_two\_sided/1}] \\
Lenient, (b) dropped & 777\src{bench/results/v3/human\_audit/audit\_report.json\#lenient/H1\_with\_human\_grades\_ungraded\_dropped/questions} & 77.5\%\src{bench/results/v3/human\_audit/audit\_report.json\#lenient/H1\_with\_human\_grades\_ungraded\_dropped/acc\_t0r} & 79.9\%\src{bench/results/v3/human\_audit/audit\_report.json\#lenient/H1\_with\_human\_grades\_ungraded\_dropped/acc\_engram} & $-$2.4\src{bench/results/v3/human\_audit/audit\_report.json\#lenient/H1\_with\_human\_grades\_ungraded\_dropped/d\_bar} & $-$4.6\src{bench/results/v3/human\_audit/audit\_report.json\#lenient/H1\_with\_human\_grades\_ungraded\_dropped/lower\_bound\_95\_one\_sided} & [$-$5.0\src{bench/results/v3/human\_audit/audit\_report.json\#lenient/H1\_with\_human\_grades\_ungraded\_dropped/ci95\_two\_sided/0},~+0.09\src{bench/results/v3/human\_audit/audit\_report.json\#lenient/H1\_with\_human\_grades\_ungraded\_dropped/ci95\_two\_sided/1}] \\
\bottomrule
\end{tabular}
\end{table}
No conclusion changes. H1 is non-inferior under all four. The worst bound is $-$4.7\src{bench/results/v3/human\_audit/audit\_report.json\#lenient/H1\_with\_human\_grades/lower\_bound\_95\_one\_sided} under (a) and $-$4.6\src{bench/results/v3/human\_audit/audit\_report.json\#lenient/H1\_with\_human\_grades\_ungraded\_dropped/lower\_bound\_95\_one\_sided} under (b). One statement depends on the choice. Under lenient grading with (a), the two-sided interval lies just below zero, so by that grading engram v2 is more accurate. With (b), the interval reaches +0.09\src{bench/results/v3/human\_audit/audit\_report.json\#lenient/H1\_with\_human\_grades\_ungraded\_dropped/ci95\_two\_sided/1}, so the difference is not detected.

The grades, the key and the analysis are in \texttt{bench/\allowbreak{}results/\allowbreak{}v3/\allowbreak{}human\_\allowbreak{}audit/\allowbreak{}} and \texttt{bench/\allowbreak{}v3\_\allowbreak{}human\_\allowbreak{}audit.py}.

\FloatBarrier
\section{Shortlist recall}\label{app:D}
For every scored question of the nine held-out conversations, the 30\src{docs/V3\_PLAN.md §1, §2, §4}-turn cosine shortlist (shared by Turns + cosine and Turns + Jev) and the turns Turns + Jev's rerank keeps were rebuilt from the frozen stores through the call cache. Each turn's id is its LoCoMo dialogue id, so the question's evidence ids can be located. Reported: the share of questions with all, and with at least one, evidence turn in the shortlist, and, among questions with an evidence turn in the shortlist, the share where the rerank keeps none. Recall is scored on raw turns only \citep{samerank2026}. The five fresh conversations (76.9\%\src{bench/results/v3/shortlist\_recall.json\#fresh\_five/all/shortlist\_recall\_all} all-evidence recall, 10.4\%\src{bench/results/v3/shortlist\_recall.json\#fresh\_five/all/rerank\_drops\_all\_shortlisted\_evidence} dropped by the rerank) and the four exploratory ones (76.9\%\src{bench/results/v3/shortlist\_recall.json\#exploratory\_four/all/shortlist\_recall\_all}, 10.3\%\src{bench/results/v3/shortlist\_recall.json\#exploratory\_four/all/rerank\_drops\_all\_shortlisted\_evidence}) agree.

\begin{table}[!htbp]
\centering
\footnotesize\hyphenpenalty=10000\exhyphenpenalty=10000\setlength{\tabcolsep}{6.0pt}

\begin{tabular}{@{}>{\raggedright\arraybackslash}p{183.6pt}>{\raggedleft\arraybackslash}p{63.7pt}>{\raggedleft\arraybackslash}p{59.0pt}>{\raggedleft\arraybackslash}p{47.0pt}>{\raggedleft\arraybackslash}p{53.8pt}@{}}
\toprule
\textbf{Category} & \textbf{Questions} & \textbf{All evidence in shortlist} & \textbf{At least one} & \textbf{Rerank keeps none} \\
\midrule
Multi-hop & 250\src{bench/results/v3/shortlist\_recall.json\#all\_nine/by\_category/multi-hop/questions} & 42.8\%\src{bench/results/v3/shortlist\_recall.json\#all\_nine/by\_category/multi-hop/shortlist\_recall\_all} & 88.4\%\src{bench/results/v3/shortlist\_recall.json\#all\_nine/by\_category/multi-hop/shortlist\_recall\_any} & 5.9\%\src{bench/results/v3/shortlist\_recall.json\#all\_nine/by\_category/multi-hop/rerank\_drops\_all\_shortlisted\_evidence} \\
Temporal & 284\src{bench/results/v3/shortlist\_recall.json\#all\_nine/by\_category/temporal/questions} & 83.5\%\src{bench/results/v3/shortlist\_recall.json\#all\_nine/by\_category/temporal/shortlist\_recall\_all} & 88.4\%\src{bench/results/v3/shortlist\_recall.json\#all\_nine/by\_category/temporal/shortlist\_recall\_any} & 22.7\%\src{bench/results/v3/shortlist\_recall.json\#all\_nine/by\_category/temporal/rerank\_drops\_all\_shortlisted\_evidence} \\
Open-domain & 83\src{bench/results/v3/shortlist\_recall.json\#all\_nine/by\_category/open-domain/questions} & 42.0\%\src{bench/results/v3/shortlist\_recall.json\#all\_nine/by\_category/open-domain/shortlist\_recall\_all} & 63.0\%\src{bench/results/v3/shortlist\_recall.json\#all\_nine/by\_category/open-domain/shortlist\_recall\_any} & 31.4\%\src{bench/results/v3/shortlist\_recall.json\#all\_nine/by\_category/open-domain/rerank\_drops\_all\_shortlisted\_evidence} \\
Single-hop & 771\src{bench/results/v3/shortlist\_recall.json\#all\_nine/by\_category/single-hop/questions} & 89.2\%\src{bench/results/v3/shortlist\_recall.json\#all\_nine/by\_category/single-hop/shortlist\_recall\_all} & 91.2\%\src{bench/results/v3/shortlist\_recall.json\#all\_nine/by\_category/single-hop/shortlist\_recall\_any} & 5.8\%\src{bench/results/v3/shortlist\_recall.json\#all\_nine/by\_category/single-hop/rerank\_drops\_all\_shortlisted\_evidence} \\
All & 1,388\src{bench/results/v3/shortlist\_recall.json\#all\_nine/all/questions} & 76.9\%\src{bench/results/v3/shortlist\_recall.json\#all\_nine/all/shortlist\_recall\_all} & 88.5\%\src{bench/results/v3/shortlist\_recall.json\#all\_nine/all/shortlist\_recall\_any} & 10.4\%\src{bench/results/v3/shortlist\_recall.json\#all\_nine/all/rerank\_drops\_all\_shortlisted\_evidence} \\
\bottomrule
\end{tabular}
\end{table}
\FloatBarrier
\section{Turns + Jev (wide), post-hoc exploratory}\label{app:E}
Designed after the registered results were seen, tested once on the 778\src{bench/results/v3/batch\_b\_report.json\#H1/questions} questions of the five held-out conversations, outside the Holm family, in its own ledger (\$0.31\src{bench/results/v3\_posthoc/report.json\#ledger/openai} OpenAI and \$0.58\src{bench/results/v3\_posthoc/report.json\#ledger/jev} Jev). Design: Turns + Jev's store; a 150\src{bench/run.py\#ARMS/lean\_t0r\_wide/flags/retrieval\_shortlist}-turn cosine shortlist; Jev's relevance question on every shortlisted turn, thirty per request; the top k by Jev's probability, with no cut-off, floor or expansion; $k{=}$47\src{bench/results/v3\_posthoc/report.json\#t0r\_wide\_k} matched to Jev-Mem at $k{=}$40 by the registered rule, saved before answering. Results: accuracy 81.5\%\src{bench/results/v3\_posthoc/report.json\#comparison/T0R-wide 47 (POST-HOC EXPLORATORY (T0R-wide; docs/V3\_PLAN.md §12))/accuracy} at 2,000\src{bench/results/v3\_posthoc/report.json\#comparison/T0R-wide 47 (POST-HOC EXPLORATORY (T0R-wide; docs/V3\_PLAN.md §12))/tokens\_mean} tokens (multi-hop 78.6\src{bench/results/v3\_posthoc/report.json\#comparison/T0R-wide 47 (POST-HOC EXPLORATORY (T0R-wide; docs/V3\_PLAN.md §12))/by\_category/multi-hop}, temporal 76.4\src{bench/results/v3\_posthoc/report.json\#comparison/T0R-wide 47 (POST-HOC EXPLORATORY (T0R-wide; docs/V3\_PLAN.md §12))/by\_category/temporal}, open-domain 64.0\src{bench/results/v3\_posthoc/report.json\#comparison/T0R-wide 47 (POST-HOC EXPLORATORY (T0R-wide; docs/V3\_PLAN.md §12))/by\_category/open-domain}, single-hop 86.5\src{bench/results/v3\_posthoc/report.json\#comparison/T0R-wide 47 (POST-HOC EXPLORATORY (T0R-wide; docs/V3\_PLAN.md §12))/by\_category/single-hop}); read cost \$0.00089\src{bench/results/v3\_posthoc/report.json\#comparison/T0R-wide 47 (POST-HOC EXPLORATORY (T0R-wide; docs/V3\_PLAN.md §12))/read\_cost\_per\_query} per query; live read latency 720\src{bench/results/v3\_posthoc/report.json\#read\_latency\_live/T0R-wide 47/p50\_ms} ms at p50 and 776\src{bench/results/v3\_posthoc/report.json\#read\_latency\_live/T0R-wide 47/p90\_ms} ms at p90. With the same recall method on the 150\src{bench/run.py\#ARMS/lean\_t0r\_wide/flags/retrieval\_shortlist}-turn shortlist, all evidence was in the shortlist for 91.9\%\src{bench/results/v3\_posthoc/report.json\#shortlist\_recall/all/shortlist\_recall\_all} of questions and at least one turn for 97.9\%\src{bench/results/v3\_posthoc/report.json\#shortlist\_recall/all/shortlist\_recall\_any}, and the top k kept none of the shortlisted evidence for 0.9\%\src{bench/results/v3\_posthoc/report.json\#shortlist\_recall/all/top\_k\_drops\_all\_shortlisted\_evidence}.

\FloatBarrier
\section{Prompts and the Jev question}\label{app:F}
The answer prompt is mem0's LoCoMo answer prompt adapted to one memory list, and the judge is mem0's LoCoMo accuracy prompt (\texttt{bench/\allowbreak{}locomo\_\allowbreak{}subset.py}, \texttt{ANSWER\_\allowbreak{}PROMPT} and \texttt{ACCURACY\_\allowbreak{}PROMPT}). LongMemEval questions carry their question date in the question slot. Turns + Jev's rerank asks Jev one yes/no question per shortlisted turn (\texttt{src/\allowbreak{}engram/\allowbreak{}decide/\allowbreak{}questions.py}, \texttt{RELEVANT\_\allowbreak{}TO\_\allowbreak{}QUERY}):

\begin{itemize}
\item instructions: "Does \texttt{memory} help answer \texttt{query}?"
\item true: ``It states or directly implies part of the answer.''
\item false: ``It is off-topic or only shares a keyword.''
\end{itemize}
\FloatBarrier
\section{The mem0 serving incident}\label{app:G}
mem0 2.1.0 sends its OpenAI LLM calls to OpenRouter whenever an OpenRouter key is present in the environment, without warning (\texttt{mem0/\allowbreak{}llms/\allowbreak{}openai.py}). A key added for the second answer model therefore routed all 3,122\src{docs/V3\_PLAN.md §12 (mem0 via OpenRouter, resolved): provider fields of the cached responses} of mem0's extraction calls through OpenRouter, which served 1,599\src{docs/V3\_PLAN.md §12 (mem0 via OpenRouter, resolved)} of them by OpenAI and 1,523\src{docs/V3\_PLAN.md §12 (mem0 via OpenRouter, resolved)} by Azure, all as gpt-4o-mini, the registered model. OpenRouter billed \$2.42\src{bench/results/v3/spend.jsonl\#correction/openrouter\_outside\_cap}; at OpenAI list price the same calls cost \$4.12\src{bench/results/v3/spend.jsonl\#correction/spend/openai (negated)}. S3's difference is far larger than a serving difference could explain, and S3 is reported with this caveat. Before any later run, mem0 was pinned to the OpenAI endpoint and the spend guard was extended to OpenRouter. Anyone benchmarking mem0 2.1.0 with an OpenRouter key in their environment will get routed calls without warning.

\FloatBarrier
\section{Cost accounting}\label{app:H}
Every system's gpt-4o-mini, embedding and Jev calls are priced at list price (gpt-4o-mini \$0.15\src{src/engram/llm/openai.py\#PRICES/gpt-4o-mini/0 (USD per million)} and \$0.60\src{src/engram/llm/openai.py\#PRICES/gpt-4o-mini/1 (USD per million)} per million input and output tokens; Jev \$0.042\src{src/engram/config.py\#JEV\_PRICE\_PER\_INPUT\_TOKEN (per million)} per million input tokens), so no system looks cheaper because of a discount the others did not get. Billed amounts are reported separately: the registered ledger records \$23.16\src{bench/results/v3/spend.jsonl (sum of spend/openai)} of OpenAI, \$5.55\src{bench/results/v3/spend.jsonl (sum of spend/jev)} of Jev and \$0.31\src{bench/results/v3/spend.jsonl (sum of spend/openrouter)} of OpenRouter for the second answer model (at \$0.10\src{docs/V3\_PLAN.md §1, §2, §4} and \$0.32\src{docs/V3\_PLAN.md §1, §2, §4} per million input and output tokens), plus mem0's \$2.42\src{bench/results/v3/spend.jsonl\#correction/openrouter\_outside\_cap} through OpenRouter. Write-cost ratios use the held-out measurements; read costs are per query; totals per question state their reads-per-write assumption (Figure~\ref{fig:8}).

\FloatBarrier
\section{Reproduction}\label{app:I}
Each table and figure is rebuilt from the committed result files, with no API calls:

\begin{itemize}
\item numbers: \texttt{uv run -{}-extra bench python paper\_\allowbreak{}v3/\allowbreak{}make\_\allowbreak{}numbers.py}
\item figures: \texttt{uv run -{}-with matplotlib -{}-with pymupdf python paper\_\allowbreak{}v3/\allowbreak{}figures.py} (Figures~\ref{fig:1} and \ref{fig:2} and the \hyperref[app:J]{Appendix~J} figure are TikZ, from \texttt{paper\_\allowbreak{}v3/\allowbreak{}diagram.py}, compiled with pdflatex)
\item the worked examples of Figure~\ref{fig:2} and \hyperref[app:J]{Appendix~J}: \texttt{bench/\allowbreak{}v3\_\allowbreak{}worked\_\allowbreak{}example.py}, which replays both read paths from the frozen stores through the call cache opened read-only (a cache miss stops it, so it cannot call an API)
\item paper: \texttt{make -C paper\_\allowbreak{}v3 paper} (renders \texttt{main.md} and \texttt{main.tex}, builds the PDF, runs \texttt{paper\_\allowbreak{}v3/\allowbreak{}check.py})
\item reports behind the tables: \texttt{bench/\allowbreak{}v3\_\allowbreak{}report.py} (Batches A--C), \texttt{bench/\allowbreak{}v3\_\allowbreak{}human\_\allowbreak{}audit.py}, \texttt{bench/\allowbreak{}v3\_\allowbreak{}shortlist\_\allowbreak{}recall.py}, \texttt{bench/\allowbreak{}v3\_\allowbreak{}latency.py}, \texttt{bench/\allowbreak{}v3\_\allowbreak{}second\_\allowbreak{}model.py}, \texttt{bench/\allowbreak{}v3\_\allowbreak{}posthoc.py}
\end{itemize}
The runs themselves are \texttt{bench/\allowbreak{}run.py} with \texttt{-{}-study v3}, \texttt{bench/\allowbreak{}jevmem\_\allowbreak{}run.py} and \texttt{bench/\allowbreak{}v3\_\allowbreak{}batch\_\allowbreak{}c.py}, from tag \texttt{v3-frozen} onward, as recorded in the plan.

\FloatBarrier
\section{A counter-example}\label{app:J}
\begin{figure*}[tbp]
\centering
\includegraphics[width=\textwidth]{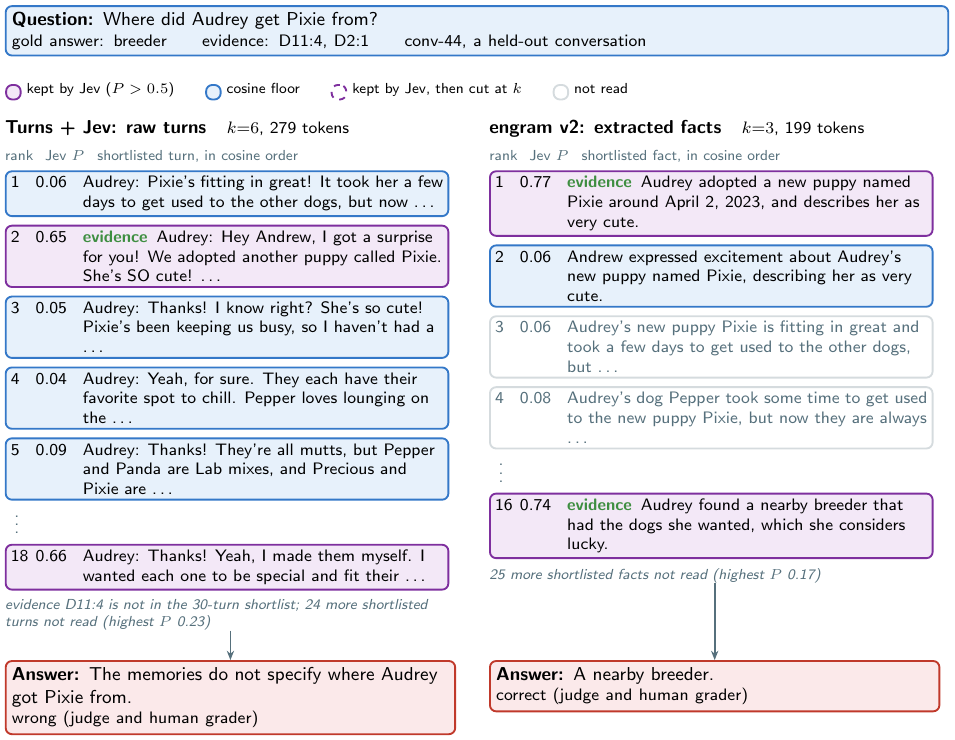}
\caption{The counter-example of \hyperref[app:J]{Appendix~J}, drawn as Figure~\ref{fig:2} (both contexts match the recorded token counts: Turns + Jev 279\src{bench/results/v3/counter\_example.json\#t0r/recorded\_tokens}, engram v2 199\src{bench/results/v3/counter\_example.json\#engram\_v2/recorded\_tokens}). The evidence turn for ``a breeder'' is not in Turns + Jev's 30\src{bench/results/v3/counter\_example.json\#t0r/shortlist (length)}-turn shortlist; engram v2's fact from it is, and Jev keeps it. An illustration chosen by the rule above, not evidence.}\label{fig:9}
\end{figure*}

Figure~\ref{fig:2} shows a question where selection wins. Figure~\ref{fig:9} shows the opposite. It was chosen by the same kind of rule and replayed the same way, with no API call. The question must:

\begin{enumerate}
\item be an H1 question that the judge and the human grader both scored correct for engram v2 and wrong for Turns + Jev (44\src{bench/results/v3/counter\_example.json\#candidates\_meeting\_rule} questions);
\item have replayed contexts that match the recorded ones;
\item come first by conversation and question index among those left.
\end{enumerate}
The question asks where Audrey got Pixie. The answer, a breeder, is in a turn that does not name Pixie (``I got lucky finding a breeder nearby that has the dogs I wanted''). That turn did not reach Turns + Jev's 30\src{bench/results/v3/counter\_example.json\#t0r/shortlist (length)}-turn cosine shortlist, which turns about Pixie fill. Jev kept the turn about her adoption (P = 0.65\src{bench/results/v3/counter\_example.json\#t0r/shortlist/1/p\_relevant}) and one unrelated turn (P = 0.66\src{bench/results/v3/counter\_example.json\#t0r/shortlist/17/p\_relevant}). Turns + Jev answered that the memories do not say. engram v2's extraction had rewritten the turn as a fact: ``Audrey found a nearby breeder that had the dogs she wanted''. That fact ranked 16\src{bench/results/v3/counter\_example.json\#engram\_v2/shortlist/15/rank}th in its fact shortlist. Jev kept it (P = 0.74\src{bench/results/v3/counter\_example.json\#engram\_v2/shortlist/15/p\_relevant}), and it reached the answer model at $k{=}$3. This is the shortlist-miss failure of \cref{sec:5.6}: a fact extracted from a turn can be retrieved when the turn itself is not.

\FloatBarrier
\end{document}